\documentclass[10 pt, journal, twoside]{ieeetran}
\usepackage{amsmath,epsfig}
\usepackage{setspace}
\usepackage{bm}
\usepackage{cite}
\usepackage{color}
\usepackage{float} 
\usepackage{array}
\usepackage{bbding}
\usepackage{pifont}
\usepackage{amssymb}
\usepackage{wasysym}
\usepackage{pdfpages}
\usepackage{epstopdf}
\usepackage{verbatim}
\usepackage{threeparttable}
\usepackage[colorlinks, linkcolor=red, anchorcolor=blue, citecolor=green]{hyperref}
\usepackage{lipsum,cite,multicol,multirow,amssymb,graphicx,array,epstopdf,subfigure,makecell,color, algorithm, algorithmic,booktabs}

\ifCLASSINFOpdf
\else
\usepackage[dvips]{graphicx}
\fi
\usepackage{url}
\usepackage{graphicx}

\begin{document}
\title{Domain Adaptation for Underwater Image Enhancement}
\author{Zhengyong~Wang, Liquan~Shen, Mei~Yu, Kun~Wang, Yufei~Lin and Mai~Xu}
\markboth{Journal of \LaTeX\ Class Files,~Vol.~14, No.~8, August~2015}
{Shell \MakeLowercase{\textit{et al.}}: Bare Demo of IEEEtran.cls for IEEE Journals}
\maketitle

\begin{abstract} 
Recently, learning-based algorithms have shown impressive performance in underwater image enhancement. Most of them resort to training on synthetic data and achieve outstanding performance. However, these methods ignore the significant domain gap between the synthetic and real data (i.e., inter-domain gap), and thus the models trained on synthetic data often fail to generalize well to real underwater scenarios. Furthermore, the complex and changeable underwater environment also causes a great distribution gap among the real data itself (i.e., intra-domain gap). However, almost no research focuses on this problem and thus their techniques often produce visually unpleasing artifacts and color distortions on various real images. Motivated by these observations, we propose a novel Two-phase Underwater Domain Adaptation network (TUDA) to simultaneously minimize the inter-domain and intra-domain gap. Concretely, a new dual-alignment network is designed in the first phase, including a translation part for enhancing realism of input images, followed by an enhancement part. With performing image-level and feature-level adaptation in two parts by jointly adversarial learning, the network can better build invariance across domains and thus bridge the inter-domain gap. In the second phase, we perform an easy-hard classification of real data according to the assessed quality of enhanced images, where a rank-based underwater quality assessment method is embedded. By leveraging implicit quality information learned from rankings, this method can more accurately assess the perceptual quality of enhanced images. Using pseudo labels from the easy part, an easy-hard adaptation technique is then conducted to effectively decrease the intra-domain gap between easy and hard samples. Extensive experimental results demonstrate that our TUDA is superior to existing works in terms of both visual quality and quantitative metrics.
\end{abstract}
\IEEEpeerreviewmaketitle

\section{Introduction}
\IEEEPARstart{I}{n} the underwater, the captured images always suffer from several kinds of degradation, including blurriness, color casts and low contrast. As light travels in the water, red light, which has longer wavelength than green and blue light, is absorbed faster, and thus underwater images often appear in a typical bluish or greenish tone. Furthermore, large amounts of suspended particles often change the direction of light in the water, resulting in dim and fuzzy images. Excellent underwater image enhancement methods are expected to improve low visibility, eliminate color casts and stretch low contrast, which can effectively enhance visual quality of input images. Meanwhile, the enhanced visibility can make scenes and objects more highlighted, providing a better starting point for high-level computer vision tasks, such as object detection and recognition.

\begin{figure}[!t]
	\centering
	\centerline{\includegraphics[width=8.4cm, height=5.4 cm]{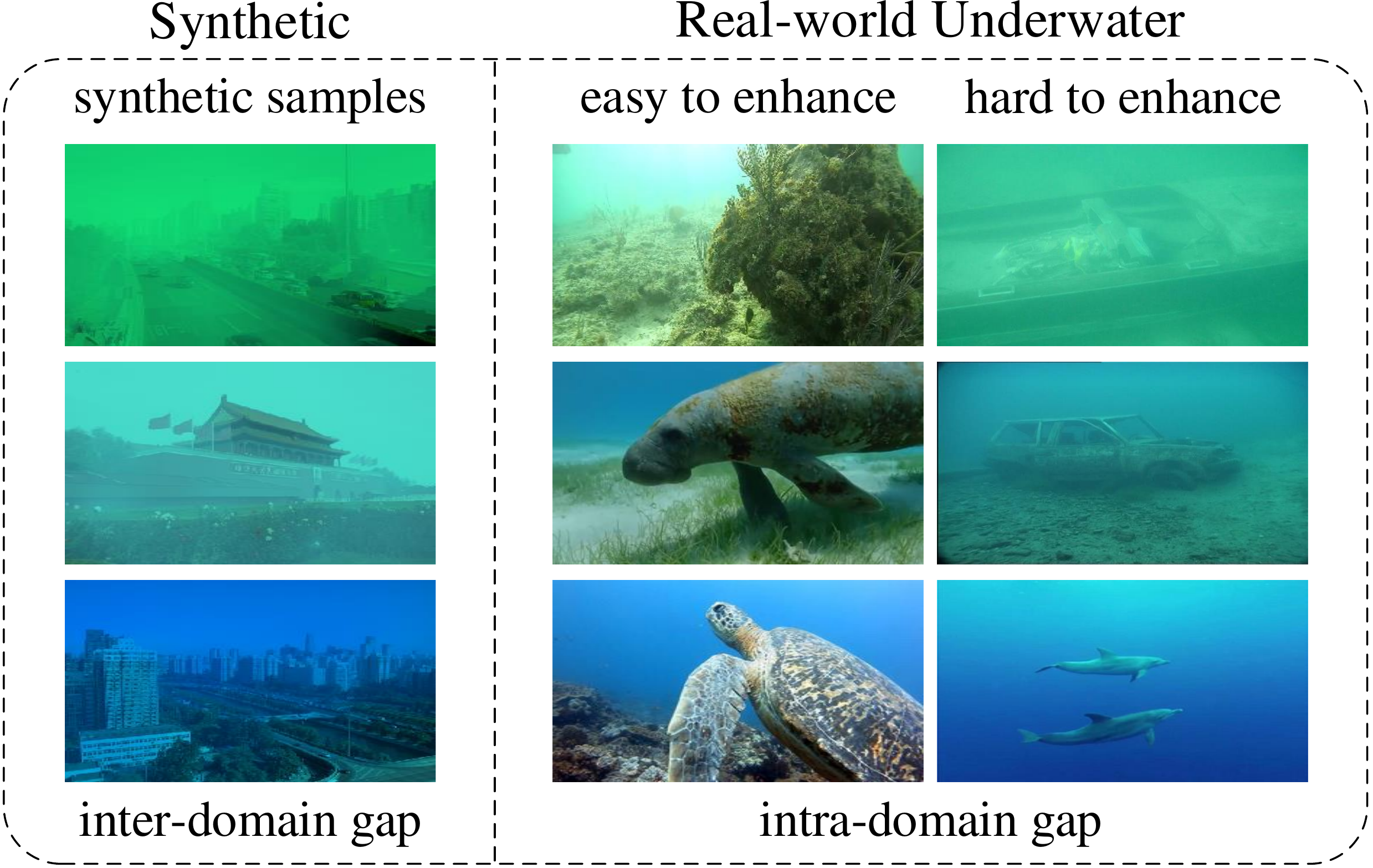}}
	\caption{Illustration of two challenges for underwater image enhancement. (1) Inter-domain gap challenge: the domain shift between the synthetic images and real images is often ignored, and thus the deep models trained on synthetic data often suffer a great performance drop in some real underwater images with different distortion distributions; (2) Intra-domain gap challenge: the complex and changeable underwater environment causes a large gap in the real-world data itself. Without considering it, deep models often produce visually unpleasing artifacts and color distortions on various real images.}
	\label{Fig_1}
\end{figure}
 
In the past decades, many algorithms have been proposed to enhance underwater images, ranging from traditional methods (image-based~\cite{Hummel1975ImageEB, Zuiderveld1994ContrastLA, Hitam2013, Ancuti2012, Fu2014, Zhang2017UnderwaterIE, Ancuti2018, Ancuti2020ColorCC} and physical-based~\cite{Chiang2012UnderwaterIE, Drews2016UnderwaterDE, Li2016UnderwaterIE, Peng2017UnderwaterIR, Berman2020UnderwaterSI, Akkaynak2019}) to learning-based methods~\cite{Li2020UnderwaterSP, Li2018WaterGANUG, Fabbri2018, Li2018EmergingFW, Li2020AnUI, Chen2019TowardsRA,Li2021}. Compared to traditional methods, learning-based methods tend to design end-to-end modules or integrate networks with physical priors to solve problems, which have better feature representation that benefits from the large data and powerful computational ability. Unfortunately, it is impractical in real-world to collect a number of real underwater images with distortion-free counterparts. Compared to real underwater data, synthetic data is much easier to be obtained. Thus, most deep methods exploit synthetic data to train the proposed models, achieving relatively promising performance. However, most of them ignore the domain shift problem from synthetic to real data, i.e., inter-domain gap, as shown in Fig.\ref{Fig_1}. These models learned from synthetic data often suffer a severe performance drop when facing some real underwater images with different distortion distributions.

\begin{figure*}[!t]
	\centering
	\centerline{\includegraphics[width=16.0cm, height=4.8cm]{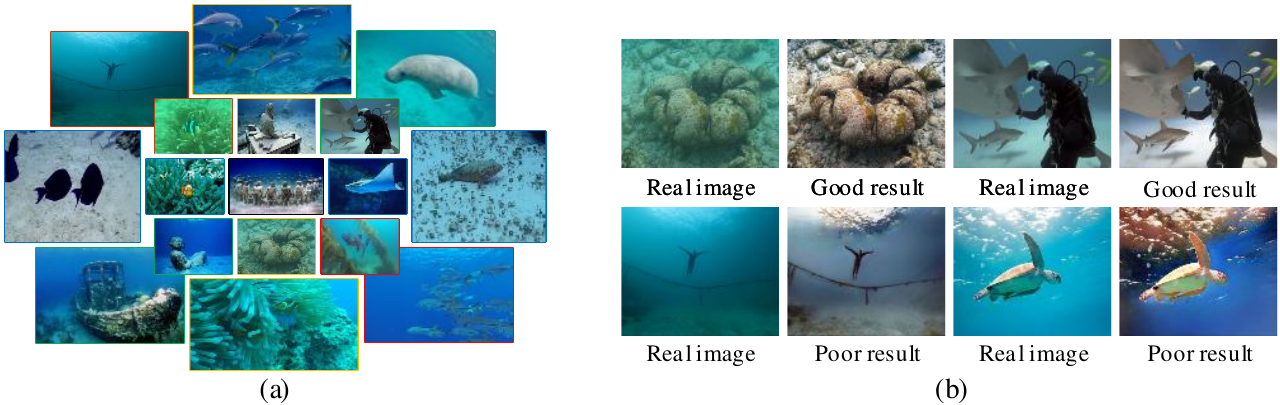}}
	\caption{(a) Examples of real-world underwater images, which have obvious different characteristics of underwater image quality degradation, e.g., color casts and blurred details. (b) Some results come from our inter-domain adaptation phase. Obviously, the results of some samples have higher perceptual quality, whereas the results of some samples suffer from local artifacts, noise, color casts and over-enhancement, etc.}
	\label{result_Easy_Hard}
\end{figure*}

Apart from this, another challenging problem in underwater image enhancement is diversity of real image distributions. Generally, the quality of images captured in the water is severely affected by many factors, such as illumination conditions, water bodies, water depth, seasonal and weather changes, etc. As shown in Fig.\ref{result_Easy_Hard}(a), these factors lead to various kinds of degradation and a large gap among real images itself, i.e., intra-domain gap. There have been rarely studies proposed to address the challenge of underwater real image itself distribution diversity. Four representative real examples and their corresponding results made by a deep model are presented in Fig.\ref{result_Easy_Hard}(b). The model shows satisfactory performance on some images  (good results). However, it cannot perform well on some images (poor results), introducing obvious local artifacts, noises and over-enhancement, etc. Obviously, without considering the intra-domain gap, it is hard for a deep model to effectively handle real underwater images with such various degradation distributions. 

Motivated by the above analysis, this paper proposes a novel \textbf{T}wo-phase \textbf{U}nderwater \textbf{D}omain \textbf{A}daptation network (TUDA) in underwater image enhancement to jointly bridge the inter-domain gap and the intra-domain gap, which consists of two phases: inter-domain adaptation and intra-domain adaptation. To be specific, a new dual-alignment network is designed in the first phase, including a translation part and an enhancement part, one for the synthetic-to-real translation and another for image enhancement. Coupled with both image-level and feature-level adaptations in an end-to-end manner, two parts can cooperative with each other for learning more domain-invariant representations to better reduce the inter-domain gap. 

In the second phase, a simple yet efficient rank-based underwater quality assessment algorithm (RUIQA) is proposed, which can better evaluate the perceptual quality of enhanced images by learning to rank. The proposed RUIQA is strongly sensitive to various artifacts and can be easily plugged in both the training and testing pipeline. Based on the assessed quality of enhanced images, we divide the real data into two categories: easy and hard samples, and get a trustworthy real image set with pseudo labels. Subsequently, using the easy-pseudo pairs and unpaired hard samples, an easy/hard domain adaptation technique is performed to close the intra-domain gap between easy and hard samples. The overview of our TUDA is presented in Fig.\ref{Framework}. To the best of our knowledge, this is the first work that successfully explores the inter-domain and intra-domain adaptation jointly in the underwater image enhancement community. The main contributions of this paper are summarized as follows:
\begin{enumerate} 
\item We propose a novel two-phase underwater domain adaptation network, called TUDA, to simultaneously reduce the inter-domain and intra-domain gap, which successfully sheds new light on future direction for enhancing underwater images. 
\item A novel dual-alignment architecture is designed in the inter-domain adaptation phase, which can effectively perform image-level and feature-level adaptations using jointly adversarial learning. Two alignment parts can improve each other, and the combination of them can better build invariance across domains and thus bridge the inter-domain gap.    
\item A rank-based underwater quality assessment method is developed in the intra-domain adaptation phase, which can effectively assess the perceptual quality of enhanced images with the help of learning to rank. From this method, we successfully perform an easy-hard classification and an easy/hard adaptation technique to significantly reduce the intra-domain gap.
\end{enumerate}

\section{Related work}
In this section, we briefly review previous related works in two aspects, i.e. underwater image enhancement and domain adaptation.
\subsection{Underwater Image Enhancement}
Underwater image enhancement approaches can be roughly categorized into three branches, i.e., image-based methods, physical-based methods and learning-based methods.

\textbf{Image-based methods}~\cite{Hummel1975ImageEB, Zuiderveld1994ContrastLA, Hitam2013, Ancuti2012, Fu2014, Zhang2017UnderwaterIE, Ancuti2018, Ancuti2020ColorCC} mainly modify pixel values of underwater images to improve visual quality, including pixel values adjustment~\cite{Hummel1975ImageEB,Zuiderveld1994ContrastLA, Hitam2013, Ancuti2020ColorCC}, retinex decomposition~\cite{Fu2014, Zhang2017UnderwaterIE} and image fusion~\cite{Ancuti2012, Ancuti2018}, etc. For example, Zhang et al.~\cite{Zhang2017UnderwaterIE} propose an extended multi-scale retinex-based underwater image enhancement method, in which the input image is processed by three steps: color correction, layer decomposition and enhancement. Ancuti et al.~\cite{Ancuti2018} propose a novel multi-scale fusion strategy, which blends a color-compensated and white-balanced version of the given image to generate a better result. Recently, based on the characteristics of severely non-uniform color spectrum distribution in underwater images, Ancuti~\textit{et al}.~\cite{Ancuti2020ColorCC} introduce a new color-channel-compensation pre-processing step in the opponent color channel to better overcome artifacts. Image-based methods can improve visual effects to some extent. However, they often fail to provide high quality results in some complex scenarios due to ignoring the domain knowledge of underwater imaging.

Most~\textbf{physical-based methods}~\cite{Chiang2012UnderwaterIE, Drews2016UnderwaterDE, Li2016UnderwaterIE, Peng2017UnderwaterIR, Berman2020UnderwaterSI, Akkaynak2019} are based on the underwater image formation model~\cite{Jaffe1990ComputerMA}, in which the background light and transmission map are estimated by some priors. The priors include underwater dark channel prior~\cite{Drews2016UnderwaterDE}, minimum information prior~\cite{Li2016UnderwaterIE}, blurriness prior~\cite{Peng2017UnderwaterIR} and color-line prior~\cite{Berman2020UnderwaterSI}, etc. For example, built on underwater image blurriness and light absorption, Peng~\textit{et al}.~\cite{Peng2017UnderwaterIR} propose an underwater image restoration method combined with a blurriness prior to estimate more accurate scene depth. Inspired by the minimal information loss principal, Li~\textit{et al}.~\cite{Li2016UnderwaterIE} estimate an optimal transmission map to restore underwater images, and exploit a histogram distribution prior to effectively improve the contrast and brightness. Recently, Berman~\textit{et al}.~\cite{Berman2020UnderwaterSI} incorporate the color-line prior and multiple spectral profiles information of different water types into the physical model, and employ the gray-world assumption theory to choose the best result, showing great performance on image dehazing. These methods can restore underwater images well in some cases. However, when the priors are invalid, undesired artifacts and color casts are still inevitable appear in some regions.

Recently, with the development of deep learning,~\textbf{learning-based methods}~\cite{Li2020UnderwaterSP, Li2018WaterGANUG, Fabbri2018, Li2018EmergingFW, Li2020AnUI, Chen2019TowardsRA,Li2021} have made significant progresses in underwater image enhancement. There are many methods improve performance by training their models on real underwater images. For example, to relax the need of paired training data, Li~\textit{et al}.~\cite{Li2018EmergingFW} develop a weakly supervised underwater color transfer model based on cycle-consistent generative adversarial network (CycleGAN) and real data to correct color. As a pioneering work, Li~\textit{et al}.~\cite{Li2020AnUI} build a real underwater image enhancement dataset, including totally 950 underwater raw images and reference images. The reference images are produced by 12 enhancement algorithms, and scored by 50 volunteers to choose the final results. With these images, Li~\textit{et al}.~\cite{Li2020AnUI} design a gate fusion network, in which three confidence maps are learned to fuse three pre-processing versions into a decent result. Recently, Li~\textit{et al}.~\cite{Li2021} develop an underwater image enhancement network in medium transmission-guided multi-color space for more robust enhancement. The methods trained on real data can produce visually pleasing results. However, they cannot restore the color and structure of objects well and tend to produce inauthentic results since the reference images are not the actual ground truths. 

There are also many algorithms to train their networks using data synthesized from Generative Adversarial Networks~\cite{Li2018WaterGANUG, Fabbri2018} or physical models~\cite{Jaffe1990ComputerMA, dudhane2020deep}. For example, combined with the knowledge of underwater imaging, Li~\textit{et al}.~\cite{Li2018WaterGANUG} design a generative adversarial network for generating realistic underwater-like images from in-air images and depth maps, and then utilize these generated data to correct color casts in a supervised manner. Fabbri~\textit{et al}.~\cite{Fabbri2018} directly employ a CycleGAN to generate paired training data, and then a fully convolutional encoder-decoder is trained to improve underwater image quality. In addition, Li~\textit{et al}.~\cite{Li2020UnderwaterSP} propose to synthesize ten types of underwater images based on an underwater image formation model and some scene parameters. With the synthetic data, Li~\textit{et al}.~\cite{Li2020UnderwaterSP} develop an end-to-end model to directly recover the clear underwater latent image first, and then conduct a post-processing to improve subjective visual effects. Dudhane~\textit{et al}.~\cite{dudhane2020deep} improve the work of ~\cite{Li2020UnderwaterSP} by introducing the object blurriness and color shift components to synthesize more accurate underwater-like data. 

Synthesis data can simulate different underwater types and degradation levels, and has the corresponding reference images as guidance for network training. However, due to the certain domain discrepancy between synthetic and real-world data, deep models trained on synthetic data often fail to generalize well on real underwater scenarios.

\begin{figure*}[!t]
	\centering
	\centerline{\includegraphics[scale = 0.54]{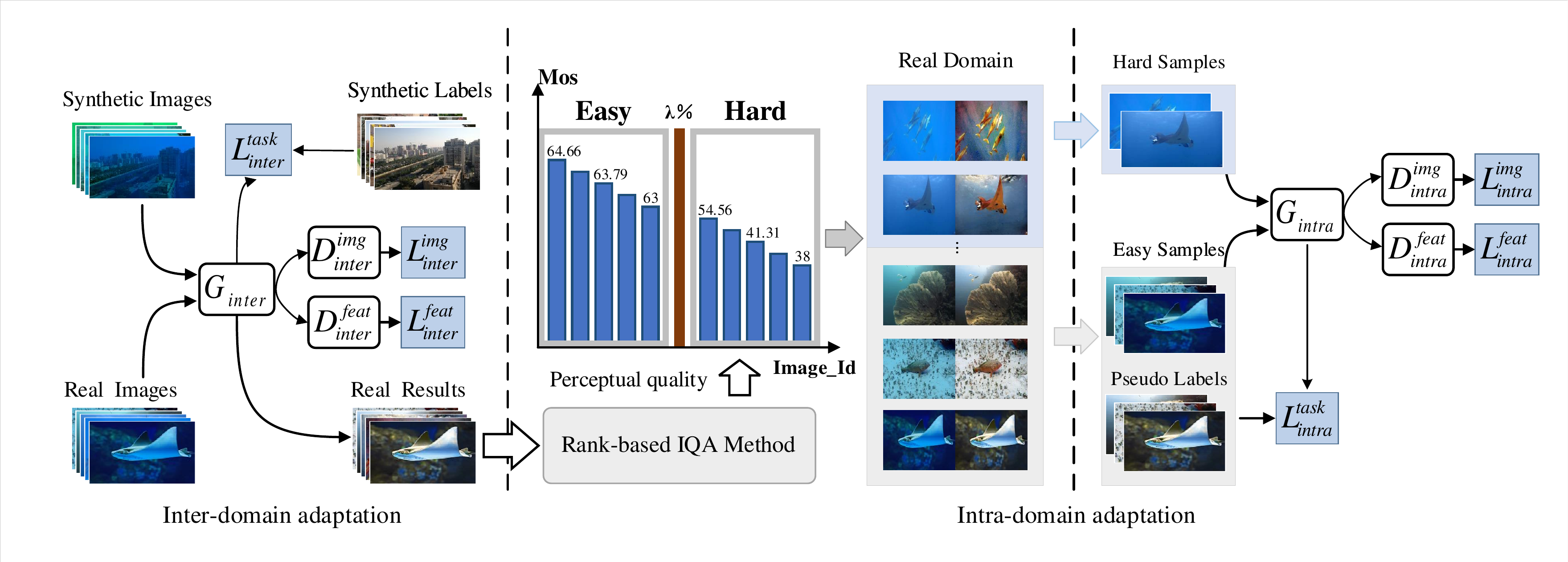}} 
	\caption{Illustration of our proposed TUDA, which consists of two phases, inter-domain adaptation and intra-domain adaptation. In the inter-domain adaptation phase, $G_{{inter}}$ can effectively reduce the distribution discrepancy between synthetic and real images using the image-level and feature-level discriminator $D_{{inter}}^{{img}}$ and $D_{{inter}}^{{feat}}$. Details are introduced in Section~\ref{section:me}. In the intra-domain adaptation phase, a rank-based underwater image quality assessment method (IQA) is first presented to separate all real data into easy and hard samples, where $\lambda$ is the ratio of real-world images assigned into the easy samples. Then, using the trustworthy easy set with generated precise pseudo labels, we can powerfully close the intra-domain gap between easy and hard samples with the help of $G_{{intra}}$,  $D_{{intra}}^{{img}}$ and $D_{{intra}}^{{feat}}$.}
	\label{Framework}
\end{figure*}

\subsection{Domain Adaptation}
Domain adaptation has been extensively explored recently, which aims to reduce the distribution gap between two different domains, and can be performed at the image level or feature level. To the best of our knowledge, domain adaptation is seldom systematic studied in underwater image enhancement field. However, it has a wide range of applications in other fields such as image hazing~\cite{Shao2020}, semantic segmentation~\cite{Pan2020, Shin2020TwoPhasePL} and depth prediction~\cite{Zheng2018, Zhao2019}, etc. For example, Shao~\textit{et al}.~\cite{Shao2020} propose a domain adaptation for single image dehazing based on CycleGAN, in which a new bidirectional translation network is design to reduce the gap between synthetic and real images by jointly synthetic-to-real and real-to-synthetic image-level adaptations. Zhao~\textit{et al}.~\cite{Zhao2019} propose a novel geometry-aware symmetric domain adaptation framework to explore the labels in the synthetic data and epipolar geometry in the real data jointly for better bridge the gap between synthetic and real domains, and thus generate high-quality depth maps.

More recently, Pan~\textit{et al}~\cite{Pan2020} propose an unsupervised intra-domain adaptation through self-supervision for semantic segmentation. To obtain extra performance gains, the authors first train the model using the inter-domain adaptation from existing approaches, and decompose the target domain in two small subdomains based on the mean value of entropy maps from the predicted segmentation maps. Then an alignment on entropy maps for both subdomains are conducted to further reduce the intra-domain gap. Inspired by this work, the concepts of inter- and intra-domain are introduced to underwater image enhancement. In this paper, we propose a different domain adaptation method, in which a new dual-alignment network used for inter-domain adaptation and a novel underwater image quality assessment algorithm used for intra-domain adaptation are proposed. The detail architectures of the proposed method are introduced in the following sections.

\section{Proposed Method}
Given a set of synthetic images $X_{S}=\left\{x_{s}, y_{s}\right\}$ and a real underwater image set $X_{R}=\left\{x_{r}\right\}$, we aim to reduce the inter-domain gap between the synthetic and real data and the intra-domain gap among the real data itself. A novel two-phase underwater domain adaptation network is proposed, which consists of two parts: inter-domain adaptation and intra-domain adaptation. As shown in Fig.\ref{Framework}, in the inter-domain phase, a new dual-alignment network $G_{{inter}}$ is developed to jointly perform image-level and feature-level alignment, including an image translation part $G_{{inter}}^{T}$ and an image enhancement part $G_{{inter}}^{E}$. The former is used for learning a more robust transformation of synthetic to real underwater images, and the latter is used for performing image enhancement using both translated and real images. Details are introduced in Section~\ref{section:inter}. From this adaptation, a rank-based underwater quality assessment method (i.e., RUIQA) is designed to evaluate the perceptual quality of the enhanced images. Based on these predicted quality scores, we separate the real underwater raw images into easy and hard samples ($X_{E}=\left\{x_{e}, y_{e}\right\}$ and $X_{H}=\left\{x_{h}\right\}$), and then conduct the intra-domain adaptation similar to inter-domain adaptation. Details of this phase are described in Section~\ref{section:intra}.
\label{section:me}

\begin{figure*}[!t]
	\centering
	\centerline{\includegraphics[width=17.0cm, height=5.6cm]{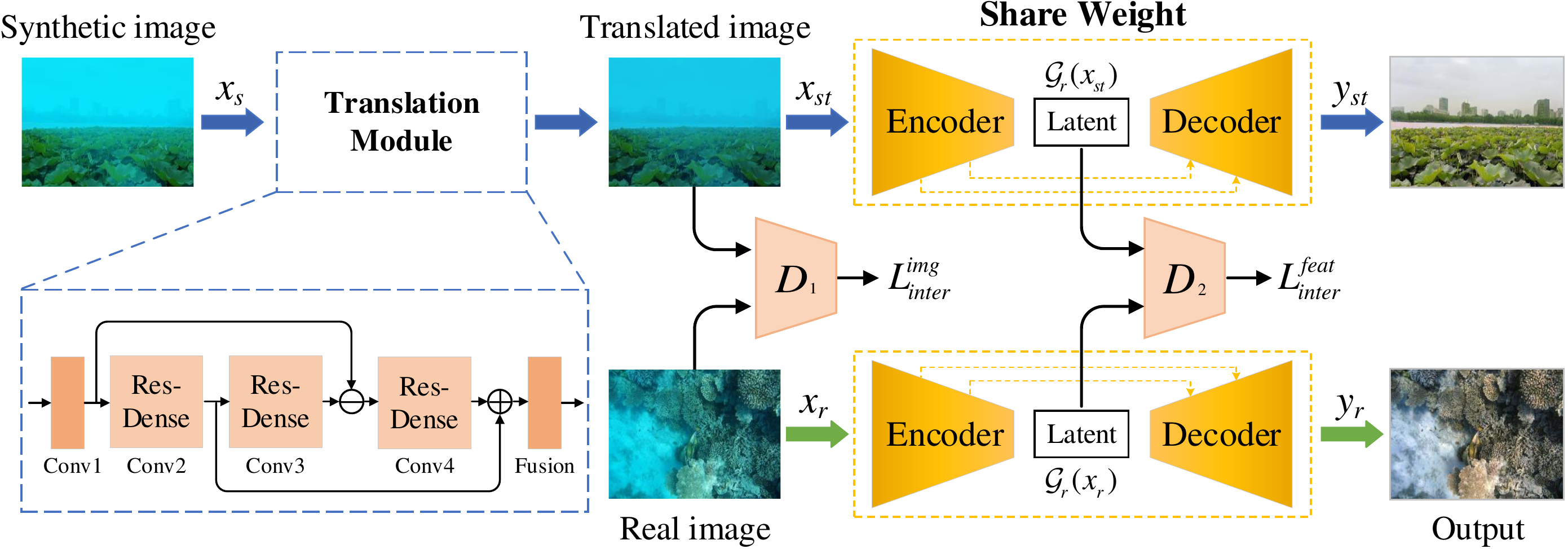}} 
	\caption{Illustration of our dual-alignment network proposed in the inter-domain adaptation phase, trained on synthetic underwater image pairs and unpaired real images, which consists of an image translation part for enhancing realism of input images, followed by an image enhancement part. They are cooperatively performed image-level and feature-level alignments and trained end-to-end in an adversarial learning manner.}
	\label{inter}
\end{figure*}

\subsection{$\text{1}^{{st}} \text{ phase:}$ Inter-domain Adaptation }
Our proposed dual-alignment network aims to reduce the inter-domain adaptation gap between the synthetic and real data domain in both image level and feature level, as shown in Fig.\ref{inter}. The proposed network composes of two parts: an image translation module $G_{{inter}}^{T}$ for enhancing realism of input images, followed by an enhancement module $G_{{inter}}^{E}$. $G_{{inter}}^{T}$ takes synthetic samples and their corresponding ground truth labels ($x_{s}, y_{s}$) as inputs, and generates translated images $x_{st}$, i.e., $x_{st}=G_{{inter}}^{T}\left(x_{s}\right)$. The translated images $x_{st}$ are expected as possible with similar distribution of real images $x_{r}$. Meanwhile, the discriminator $D_{{inter}}^{img}$ is encouraged to identify the difference between $x_{st}$ and $x_{r}$. To stabilize the gradients and improve performance, the WGAN-GP adversarial loss~\cite{Gulrajani2017} is adopted to perform image-level alignment, set as:
\begin{equation}\label{eq1}
\begin{aligned}
L_{{inter}}^{img} &=\mathbb{E}_{x_{st}}\left[D_{{inter}}^{img}\left(x_{st}\right)\right]-\mathbb{E}_{x_{r}}\left[D_{{inter}}^{img}\left(x_{r}\right)\right] \\
&\left.+\lambda_{img} \mathbb{E}_{\hat{I}}\left(\left\|\nabla_{\hat{I}} D_{{inter}}^{img}(\hat{I})\right\|_{2}-1\right)^{2}\right]
\end{aligned}
\end{equation}
where $\hat{I}$ represents the sampling distribution which is sampled uniformly from $x_{r}$ and $x_{st}$, and $\lambda_{img}$ is the penalty parameter, in our works, $\lambda_{img}=10$.
\label{section:inter}

Color cast is one of the main characteristics of underwater images, which generally can be divided into three tones: blue, green and blue-green~\cite{Liu2020RealWorldUE}. Inspired by this, the synthetic and real images are divided into three color tone subsets according to the average value of the blue (b) channel in the CIElab color space. When the synthetic images and the real images are in the same color tone, the synthetic-to-realistic translation can be accomplished, which greatly speeds up the convergence of the model. In addition, intuitively, the gap between the synthetic and real data mainly comes from low-level differences, such as color and texture. Thus, the translated images $x_{st}$ should be retained the same semantic content as $x_{s}$, but with a different appearance. Therefore, a semantic content loss component is incorporated along with the adversarial loss, set as:
\begin{equation}\label{eq2}
L_{{inter}}^{{cont}}=w_{k} \sum_{k \in L_{c}}\left\|\phi_{k}\left(x_{s}\right)-\phi_{k}\left(x_{s t}\right)\right\|_{1}
\end{equation}
where $\phi_{k}(\cdot)$ is the $k$th-layer feature extractor of the VGG-19 network pretrained on ImageNet, $L_{c}$ is the set of layers, including conv1-1, conv2-1, conv3-1, conv4-1 and conv5-1. $w_{k}$ denotes the weight of the $k$th-layer, set as $\frac{1}{32}, \frac{1}{16}, \frac{1}{8}, \frac{1}{4}, 1.0$ in our experiments.

After the synthetic images $x_{s}$ are translated, the generated realistic images $x_{st}$ can be obtained. The paired translated data ($x_{st}, y_{s}$) is utilized to train the enhancement network $G_{{inter}}^{E}$. $G_{{inter}}^{E}$ is trained in a supervised way, including a content loss and a perceptual loss, set as:
\begin{equation}\label{eq3}
\begin{aligned}
L_{{inter}}^{{task}}=a*\left\|y_{s}-y_{st}\right\|_{1}+ b*\sum_{k \in L_{c}}\left\|\phi_{k}\left(y_{s}\right)-\phi_{k}\left(y_{st}\right)\right\|_{1}
\end{aligned}
\end{equation}
where $y_{st}$ is the output of the enhancement network $G_{{inter}}^{E}$, i.e., $y_{st} = G_{{inter}}^{E}(x_{st})$. The two parameters $a$ and $b$ are the weights of different loss components, set as 0.8 and 0.2, respectively.

To better minimize the inter-domain gap, a feature-level adversarial loss is also introduced into the enhancement part, set as:
\begin{equation}\label{eq4}
\begin{aligned}
L_{{inter}}^{feat} &=\mathbb{E}_{x_{st}}\left[D_{{inter}}^{feat}\left(\mathcal{G}_{r}\left(x_{\mathrm{st}}\right)\right)\right]-\mathbb{E}_{x_{r}}\left[D_{{inter}}^{feat}\left(\mathcal{G}_{r}\left(x_{\mathrm{r}}\right)\right)\right] \\
&\left.+\lambda_{feat} \mathbb{E}_{\hat{I}}\left(\left\|\nabla_{\hat{I}} D_{{inter}}^{feat}(\hat{I})\right\|_{2}-1\right)^{2}\right]
\end{aligned}
\end{equation}
where $G_{{inter}}^{E}$ shares identical weights in both real and translated input pipelines and $\mathcal{G}_{r}$ is the encoder of $G_{{inter}}^{E}$. $\hat{I}$ denotes the sampling distribution sampled uniformly from $\mathcal{G}_{r}\left(x_{\mathrm{st}}\right)$ and $\mathcal{G}_{r}\left(x_{\mathrm{r}}\right)$. $\lambda_{feat}$ is the penalty parameter, set as 10 in our experiments.

With both image-level and feature-level alignments in an end-to-end manner, our dual-alignment network can better build invariance between domains and thus close the inter-domain gap. The overall loss function for the inter-domain adaptation phase is expressed as follows:
\begin{equation}\label{eq5}
\begin{array}{c}
L_{{inter}}=\lambda_{1}L_{{inter}}^{{img}}+\lambda_{2}L_{{inter}}^{{cont}}+\lambda_{3}L_{{inter}}^{{E}}+\lambda_{4}L_{{inter}}^{{feat}}
\end{array}
\end{equation}
where $\lambda_{1}$, $\lambda_{2}$, $\lambda_{3}$ and $\lambda_{4}$ are trade-off weights. In our work, they are set as 1, 100, 10 and 0.0005, respectively.

\begin{figure*}[htbp]
	\centering
	\centerline{\includegraphics[scale = 0.53]{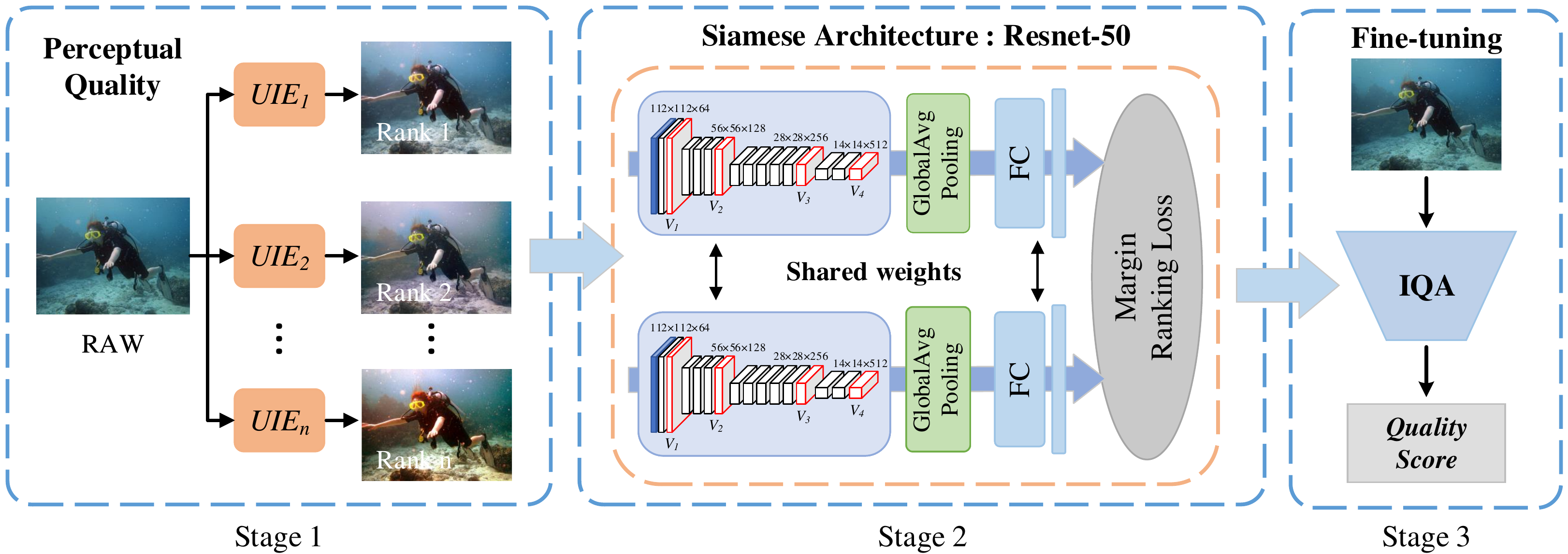}} 
	\caption{The proposed RUIQA consists of three stages, namely stage 1: build a real-world underwater ranking dataset based on their perceptual quality; stage 2: train the Siamese architecture ResNet-50 using the ranking dataset; stage 3: perform a fine-tuning technique to predict the image quality score.}
	\label{intra}
\end{figure*}

\begin{figure*}[htbp]
	\centering
	\centerline{\includegraphics[width=17.4cm, height=5.6cm]{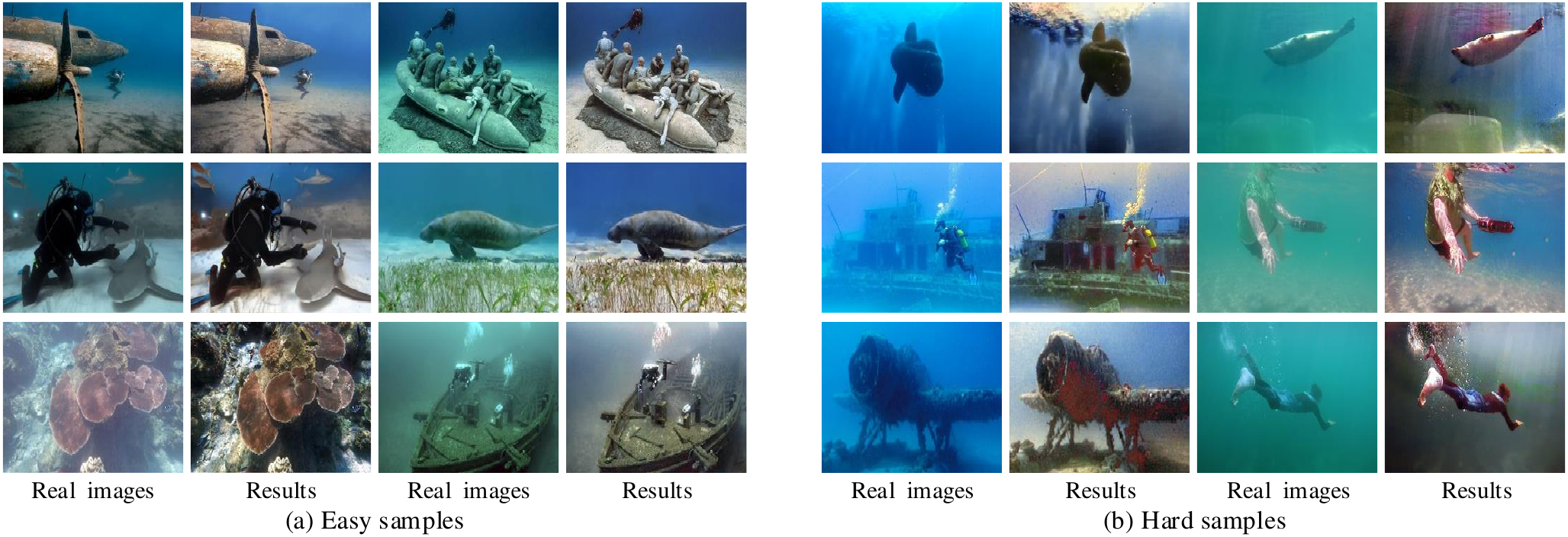}} 
	\caption{Illustration of easy and hard samples. Their results come from the same inter-domain adaptation network. Obviously, the results of easy samples have higher perceptual quality, whereas the results of hard samples suffer from local artifacts, noise, color casts and over-enhancement, etc.}
	\label{Easy_Hard}
\end{figure*}
 
\subsection{$\text {2}^{\text {nd}} \text{ phase: }$ Intra-domain Adaptation}
As mentioned above, the intra-domain gap exists among real underwater images itself, and thus a straightforward method is to divide and conquer. Some images containing a similar distribution with the training data are easy to be enhanced, called easy samples, and vice versa. Therefore, real underwater images can be separated into easy samples and hard samples according to the assessed quality of enhanced images. Enhanced results of easy samples are trustworthy, which can be used as pseudo-labels. By using easy samples and their corresponding pseudo-labels, an unsupervised way is conducted to learn easy/hard adaptation to close the intra-domain gap between easy and hard samples. To reasonably separate real underwater into easy and hard parts based on the quality of enhanced images, an effective method is required. One may attempt to use existing underwater image quality assessment methods for separating, such as UCIQE~\cite{Yang2015AnUC} and UIQM~\cite{Panetta2016HumanVisualSystemInspiredUI}. However, the experimental results in~\cite{Li2020AnUI} show that these methods cannot accurate evaluate image quality in some cases. Notably, this paper presents a novel and effective underwater quality assessment method with the help of rank information learned from rankings, which can effectively assess the quality of enhanced images, named rank-based underwater image quality assessment (RUIQA).  
\label{section:intra}

\subsubsection{Rank-based Underwater Image Quality Assessment (RUIQA)}
Existing deep IQA methods usually initialize their model parameters using the pre-trained models on the ImageNet dataset~\cite{Li2019WhichHasBetter, Su2019Blindlyassess}. Although these metrics achieve good results on ground images to some extent, the performance is unsatisfactory when facing images with various underwater distortions. In our opinion, this is mainly caused by the fact that pre-trained models capture information that is conducive to ground image processing instead of the unique prior information of underwater images, and thus they cannot easily adapt the characteristics of underwater image quality assessment tasks. 

Inspired by~\cite{Zhang2019Rankgrgan} in image super-resolution, this paper utilizes an underwater ranking dataset to train a large network to extract some ranking information by learning to rank, which is closely related to the perceptual quality. And then we fine-tune it to more accurately predict the perceptual quality of enhanced images. Differently, in~\cite{Zhang2019Rankgrgan}, a Ranker is trained to learn the behavior of perceptual metrics and then a novel rank-content loss is introduced to optimize the perceptual quality, while our method trains an underwater ranker and makes it as model initialization parameters to help assess perceptual quality. As shown in Fig.\ref{intra}, our RUIQA consists of three stages: generating rank images, training ranker and fine-tuning network. 

\textbf{Generating rank images:} A large number of underwater images are first collected from online and some public datasets~\cite{Li2020AnUI, Liu2020RealWorldUE, islam2020fast}, and then carefully selected and refined. Most of the collected images are weeded out, and about 3900 candidate images are remained. We randomly choose 800 pictures to construct an underwater ranking dataset. With the candidate underwater images, the enhanced images are generated by 8 image enhancement methods, including Fusion-12~\cite{Ancuti2012}, Fusion-18~\cite{Ancuti2018}, Two-step-based~\cite{Fu2017}, Histogram prior~\cite{Li2016UnderwaterIE}, Blurriness-based~\cite{Peng2017UnderwaterIR}, Water-Net~\cite{Li2020AnUI}, FUIE-GAN~\cite{islam2020fast} and a commercial application for enhancing underwater images (i.e., dive+). Each enhanced image is assessed with a continuous quality scale, ranging from 1 to 5. After then, the quality scale is map to a continuous score between 1 to 100. 20 volunteers are invited to conduct the evaluation in the same monitor environment. Following the work of~\cite{wu2020subjective}, the raw scores are refined by means of some standardized settings~\cite{bt2002methodology, itu1999subjective} and the Mean Opinion Score (MOS) are calculated ~\cite{sheikh2006statistical, ponomarenko2013color}, obtaining reliable subjective rating results. Our dataset and code will be publicly released on https://github.com/Underwater-Lab/TUDA.
 
\textbf{Training ranker:} With the obtained MOS values, the pair-wise images and the corresponding ranking order labels can be obtained. Meanwhile, ResNet-50~\cite{he2016deep} is employed as the Siamese network architecture to extract ranking information. The Siamese network is trained by a margin-ranking loss proposed in~\cite{Zhang2019Rankgrgan}, which is beneficial for model to learn the ranking information. After training, a single branch of the Siamese network, i.e., the pre-trained ResNet-50 model parameters on Ranking images, is extracted to initialize our backbone network. 

\textbf{Fine-tuning network:} In our RUIQA, the last global average pooling (GAP) and fully connected (FC) layer of the pre-trained ResNet-50 model are removed. To better handle distortion diversity, multi-scale features extracted from four layers (conv2-10, conv3-12, conv4-18 and the last layer) are treated as the input of four blocks. The block is composed of a 1×1 convolution, a GAP and a FC layer, mapping the multi-scale features into the corresponding perceptual quality vectors. Finally, these predicted quality vectors are regressed into a quality score. In the training phase, the network is fine-tuned by minimizing the $l_{1}$ loss between the predicted score and the MOS value label. 

Using the proposed RUIQA, the quality score of each enhanced image is predicted. The higher the value, the model is more confident with this real-world image (i.e., easy sample). This step can be named as an easy-hard classification. Some classification results are shown in Fig.~\ref{Easy_Hard}, it can be observed that the enhanced results of easy samples have higher perceptual quality and are near to the human perception. In practice, a ratio $\lambda$ is introduced to help the separation, which means the ratio of easy samples to total samples. The corresponding MOS value of the specified ratio $\lambda$ is set as a threshold to pick up easy samples and the rest images are considered as hard samples for the training. In Section~\ref{section:me2}, how to obtain the best ratio $\lambda$ is explored. It is very important for the intra-domain training and finally TUDA testing pipeline. 

\subsubsection{Easy/Hard Adaptation}
For easy samples $x_{e}$, the enhanced results $y_{e}$ are set as pseudo labels to obtain some real underwater pair data ($x_{e}, y_{e}$). By using the pair data ($x_{e}, y_{e}$), we aim to adopt an easy/hard adaptation technique to close the intra-domain gap between easy and hard samples, which is composed of an intra-domain translation part $G_{{intra}}^{T}$ and an intra-domain enhancement part $G_{{intra}}^{E}$. $G_{{intra}}^{T}$ tries to translate the easy sample $x_{e}$ to be indistinguishable from the hard images $x_{h}$. Meanwhile, a discriminator $D_{{intra}}^{{lmg}}$ aims to differentiate between the translated image $x_{et}$ and hard images $x_{h}$. This minimax game can be modeled using an adversarial loss as
\begin{equation}\label{eq6}
\begin{aligned}
L_{{intra}}^{img} &=\mathbb{E}_{x_{et}}\left[D_{{intra}}^{img}\left(x_{et}\right)\right]-\mathbb{E}_{x_{h}}\left[D_{{intra}}^{img}\left(x_{h}\right)\right] \\
&\left.+\lambda_{img} \mathbb{E}_{\hat{I}}\left(\left\|\nabla_{\hat{I}} D_{{intra}}^{img}(\hat{I})\right\|_{2}-1\right)^{2}\right]
\end{aligned}
\end{equation}
where the parameter $\lambda_{img}=10$, $\hat{I}$ represents the sampling distribution which is sampled uniformly from $x_{h}$ and $x_{et}$.

Similar to $G_{{inter}}^{T}$, an excellent translation $G_{{intra}}^{T}$ should keep the translated image $x_{et}$ “similar” in content to the original easy image $x_{e}$. Thus, semantic content loss is incorporated to better achieve content preservation, set as:
\begin{equation}\label{eq7}
L_{{intra}}^{{cont}}=w_{k} \sum_{k \in L_{c}}\left\|\phi_{k}\left(x_{e}\right)-\phi_{k}\left(x_{\text {et }}\right)\right\|_{1}
\end{equation}
where $L_{c}$ is the set of layers (conv1-1, conv2-1, conv3-1, conv4-1 and conv5-1) and  $\phi_{k}(\cdot)$ is the corresponding $k$th-layer feature map in pre-trained VGG-19 model. $w_{k}$ denotes the weight of the $k$th-layer, in our work, set as $\frac{1}{32}, \frac{1}{16}, \frac{1}{8}, \frac{1}{4}, 1.0$ respectively.

Then, the translated image $x_{et}$ is input to the intra-domain enhancement part $G_{{intra}}^{E}$, and the enhanced image $y_{et}$ is obtained. $G_{{intra}}^{E}$ is trained in a supervised manner, including a content loss and a perceptual loss, set as:
\begin{equation}\label{eq8}
\begin{aligned}
L_{{intra}}^{{task}}=c*\left\|y_{e}-y_{et}\right\|_{1}+ d*\sum_{k \in L_{c}}\left\|\phi_{k}\left(y_{e}\right)-\phi_{k}\left(y_{et}\right)\right\|_{1}
\end{aligned}
\end{equation}
where $c$ and $d$ are trade-off weights, set as 0.8 and 0.2 respectively.

To better minimize the intra-domain gap between easy and hard samples in the real-world domain, we also perform a feature-level adaptation, where a discriminator $D_{{intra}}^{{feat}}$ is introduced to align the distributions between the feature map of $x_{et}$ and $x_{h}$. The loss is defined as:
\begin{equation}\label{eq9}
\begin{aligned}
L_{{intra}}^{feat} &=\mathbb{E}_{x_{et}}\left[D_{{intra}}^{feat}\left(\mathcal{G}_{h}\left(x_{\mathrm{et}}\right)\right)\right]-\mathbb{E}_{x_{h}}\left[D_{{intra}}^{feat}\left(\mathcal{G}_{h}\left(x_{\mathrm{h}}\right)\right)\right] \\
&\left.+\lambda_{feat} \mathbb{E}_{\hat{I}}\left(\left\|\nabla_{\hat{I}} D_{{intra}}^{feat}(\hat{I})\right\|_{2}-1\right)^{2}\right]
\end{aligned}
\end{equation}
where $G_{{intra}}^{E}$ shares the same weight in both translated input and hard images pipelines and $\mathcal{G}_{h}$ is the encoder of $G_{{intra}}^{E}$. $\hat{I}$ denotes the sampling distribution sampled uniformly from $\mathcal{G}_{h}\left(x_{\mathrm{et}}\right)$ and $\mathcal{G}_{h}\left(x_{\mathrm{h}}\right)$. $\lambda_{feat}$ is the penalty factor, set as 10 in this work. $G_{{intra}}^{T}$ and $G_{{intra}}^{E}$ are trained in an end-to-end manner, and thus the full loss function is as follow:
\begin{equation}\label{eq10}
\begin{array}{c}
L_{{intra}}=\lambda_{a}L_{{intra}}^{{img}}+\lambda_{b}L_{{intra}}^{\text {img}}+\lambda_{c}L_{{intra}}^{{task}}+\lambda_{d}L_{{intra}}^{{feat}}
\end{array}
\end{equation}
where $\lambda_{a}$, $\lambda_{b}$, $\lambda_{c}$ and $\lambda_{d}$ are trade-off weights. In our work, we set them as 1, 100, 10 and 0.0005, respectively.

\begin{figure*}[!t]
	\centering
	\centerline{\includegraphics[width=16.5cm, height=2.7cm]{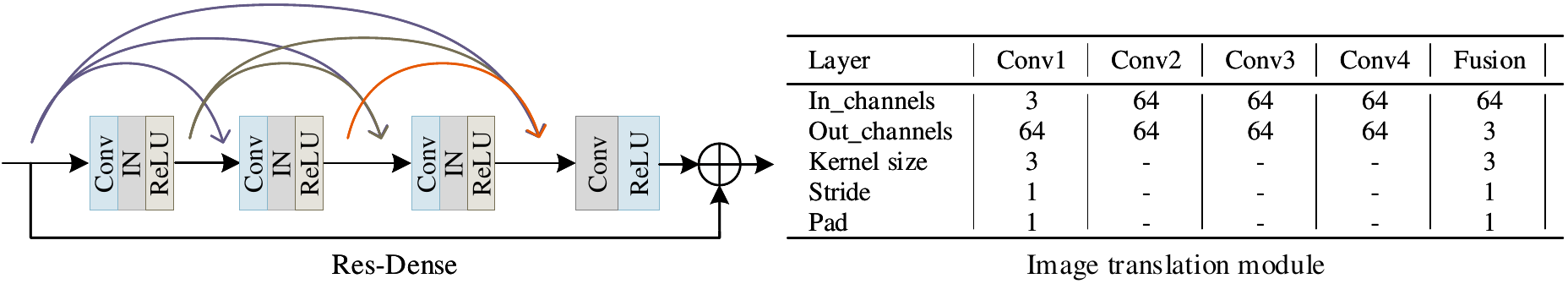}}  %
	\caption{Configurations of image translation module. "Conv1" is combined by a convolutional layer, a IN layer and a ReLU activation function. "Conv2", "Conv3", "Conv4" denote Res-Dense block. "Fusion" is combined by a convolution layer and a Tanh activation function.}
	\label{network}
\end{figure*}

\begin{figure*}[!t]
	\centering
	\centerline{\includegraphics[width=16.8cm, height=5.6cm]{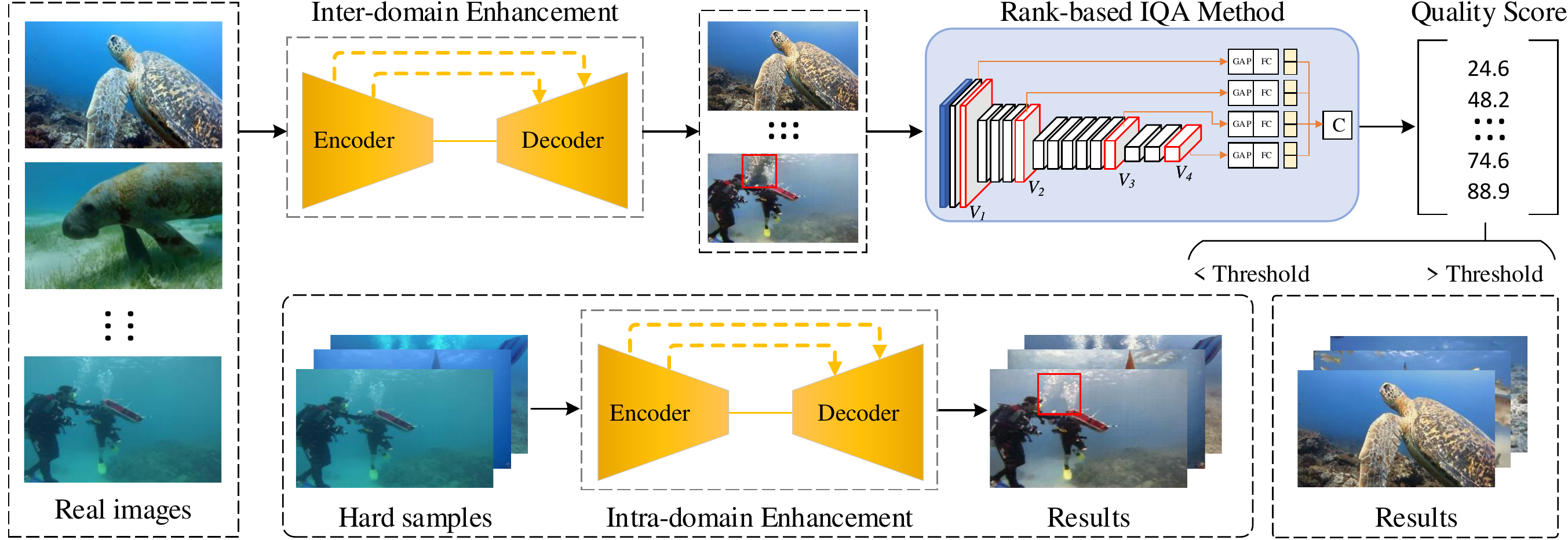}} 
	\caption{An overview of our testing pipeline. The inter-domain enhancement part first takes real underwater images as input and outputs the corresponding inter-domain enhancement results. Then, our proposed rank-based IQA method evaluates the perceived quality of the enhancement result. When the score is less than the threshold, the corresponding raw image is regarded as a hard sample, and intra-domain enhancement is performed. When the score is greater than the threshold, the result is trustworthy and output directly.}
	\label{test}
\end{figure*}

\subsection{Architecture Details}
The detail architecture of two transform modules ($G_{{inter}}^{T}$, $G_{{intra}}^{T}$) is shown in Fig.\ref{network}. The down-sampling layer is not employed in the translator for avoiding valuable information loss. For image discriminators ($D_{{inter}}^{{img}}, D_{{intra}}^{{img}}$) and feature discriminator networks ($D_{{inter}}^{{feat}}, D_{{intra}}^{{feat}}$), PatchGANs~\cite{2017ImagetoIsola} is employed, which can better locally discriminate whether image patches are real or fake. A simple network architecture (stack the dense block under the U-Net structure)\footnote[1]{{https://github.com/Underwater-Lab-SHU/ANA-SYN}} is used as our enhancement parts ($G_{{inter}}^{E}$, $G_{{intra}}^{E}$). It's worth mentioning that our test pipeline only need the enhancement parts ($G_{{inter}}^{E}$, $G_{{intra}}^{E}$) and the proposed rank-based IQA method, as shown in Fig.\ref{test}.

\section{Experiments} 
In this section, we first describe the implementation details and experiment settings of our TUDA. Then, we compare it with existing representative methods on four publicly available real underwater benchmarks. Finally, a series of ablation studies are provided to verify the advantages of each component, and the model complexity and running time are analyzed. 

\subsection{Implementation Details}
For training, a synthetic underwater image dataset is generated follow the physical model proposed in the project page of ANA-SYN. The synthetic dataset contains 9 water types\footnote[2]{{Type I, II, III, IA and IB for open ocean water and type 1C, 3C, 5C and 7C for coastal water}} defined in~\cite{Jaffe1990ComputerMA}, and each type has 1000 images which are randomly chosen from RTTS dataset~\cite{2019LiBenchmarking}. The constructed dataset is divided into two parts, 7200 ($800 \times 9$) images for training, denoted as~\textbf{Train-S7200} and 1800 ($200 \times 9$) images for testing, denoted as~\textbf{Test-S1800}. For real underwater images, as mentioned above, a large real-world underwater database including 3900 images is proposed. The database is divided two parts, 2900 images for training, denoted as~\textbf{Train-R2900} and 1000 images for testing~\textbf{Test-R1000}. All images are resized to 256 × 256 and the pixel values are normalized to $[-1,1]$. Furthermore, several data augmentation techniques are performed in the training phase, such as random rotating $90^\circ, 180^\circ, 270^\circ$ and horizontal flipping. 

Our TUDA and RUIQA are implemented in Pytorch framework and all experiments are carried out on two NVIDIA Titan V GPUs. Adam optimizer with a learning rate of $1 \times 10^{-4}$ is utilized to train $G_{{inter}}^{T}$, $G_{{inter}}^{E}$, $G_{{intra}}^{T}$ and $G_{{intra}}^{E}$. For $D_{{inter}}^{{img}}$, $D_{{inter}}^{{feat}}$, $D_{{intra}}^{{img}}$ and $D_{{intra}}^{{feat}}$, we adopt an Adam optimizer with learning rate of $2 \times 10^{-4}$ as the optimization method. Default values of $\beta_{1}$ and $\beta_{2}$ are set as 0.5 and 0.999. The batch size is set to 4. Models are trained for 200 epochs, and their learning rates decay linearly to zero in the next 100 epochs.

\subsection{Experiment Settings}
For testing, we conduct comprehensive experiments on four publicly available real-world underwater benchmarks, i.e.,~\textbf{SQUID}\footnote[3]{{The SQUID dataset contains 57 real underwater images}}~\cite{Berman2020UnderwaterSI},~\textbf{UIEB}\footnote[4]{{The UIEB dataset contains 950 real underwater images}}~\cite{Li2020AnUI},~\textbf{EUVP}\footnote[5]{{The EUVP dataset contains about 1910 real underwater images}}~\cite{islam2020fast} and~\textbf{UFO-120}\footnote[6]{{The UFO-120 dataset contains about 3255 synthetic and real images}}~\cite{2019IslamToward}. The compared algorithms include Fusion-12~\cite{Ancuti2012}, Fusion-18~\cite{Ancuti2018}, HE-Prior~\cite{Li2016UnderwaterIE}, UIBLA~\cite{Peng2017UnderwaterIR}, UGAN~\cite{Fabbri2018}, FUIE-GAN~\cite{islam2020fast} and Water-Net~\cite{Li2020AnUI}. The first four algorithms are traditional methods, while the remaining are deep-learning methods. For all the above-mentioned methods, we use the released test models and parameters to produce their results. 

 \begin{table*}[!t]
	\setlength{\abovecaptionskip}{0cm}
	\setlength{\belowcaptionskip}{-0.2cm}
	\centering
	\footnotesize
	\caption{Quantitative results (average uiqm/uciqe) of different methods on real benchmarks (Test-R1000, SQUID, UIEB, EUVP, and UFO-120). The top three results are marked in red, blue and green. '-' represents the results are not available.}
	\renewcommand{\arraystretch}{1.2}
	\begin{tabular}{c|c|c|c|c|c|c|c|c|c|c|c|c}
		\hline
		\multirow{2}{*}{Methods}&\multicolumn{6}{c|}{UCIQE$\uparrow$} &\multicolumn{6}{c}{UIQM$\uparrow$} \\
		\cline{2-13}
		&Test-R1000 &SQUID &UIEB &EUVP &UFO-120 &Avg &Test-R1000 &SQUID &UIEB &EUVP &UFO-120 &Avg \\\hline
		\hline
		\makecell[c]{Fusion-12~\cite{Ancuti2012}} &\color[rgb]{0,0.5,0}{0.607}&0.561&\color[rgb]{0,0.5,0}{0.612} &{\color{blue}{0.610}}&\color[rgb]{0,0.5,0}{0.619}&\color[rgb]{0,0.5,0}{0.613}&2.729&1.507&2.870&2.779&2.769&2.769\\
		\makecell[c]{Fusion-18~\cite{Ancuti2018}} &0.583&\color[rgb]{0,0.5,0}{0.582}&0.589 &0.588&0.596&0.591&2.879&\color[rgb]{0,0.5,0}{2.268}&\color[rgb]{0,0.5,0}{3.030}&\color[rgb]{0,0.5,0}{3.153}&\color[rgb]{0,0.5,0}{2.987}&\color[rgb]{0,0.5,0}{3.016}\\
		\makecell[c]{HE-Prior~\cite{Li2016UnderwaterIE}} &{\color{red}{0.673}}&\color[rgb]{0,0.5,0}{0.582}&{\color{red}{0.678}}&{\color{red}{0.673}}&{\color{red}{0.678}}&{\color{red}{0.675}}&-- &-- &2.637 &2.656 &2.565 &2.604\\
		\makecell[c]{UIBLA~\cite{Peng2017UnderwaterIR}} &0.581&0.500&0.595&0.594&{\color{blue}{0.624}}&0.605&2.030&--&2.262&2.053&2.061&2.183\\
		\makecell[c]{Water-Net~\cite{Li2020AnUI}} &0.576&0.537&0.584&0.578&0.598&0.587&\color[rgb]{0,0.5,0}{3.066}&2.124&2.756&2.851&2.748&2.815 \\
		\makecell[c]{FUIE-GAN~\cite{islam2020fast}} &0.548&0.488&0.560&0.556&0.582&0.566&2.777&1.791&2.949&3.079&2.954&2.952 \\
		\makecell[c]{UGAN~\cite{Fabbri2018}} &{\color{blue}{0.611}}&{\color{blue}{0.584}}&{\color{blue}{0.619}}&\color[rgb]{0,0.5,0}{0.604}&{\color{blue}{0.624}}&\color{blue}{0.615}&{\color{red}{3.130}}&{\color{red}{2.780}}&{\color{blue}{3.180}}&{\color{blue}{3.237}}&{\color{blue}{3.152}}&\color{blue}{3.172}\\
		Our TUDA &0.596&{\color{red}{0.589}}&0.605&0.596&0.608&0.602&{\color{blue}{3.125}}&{\color{blue}{2.643}}&{\color{red}{3.203}}&{\color{red}{3.313}}&{\color{red}{3.166}}&{\color{red}{3.200}}\\
		\hline\hline
	\end{tabular}
	\label{table10}
\end{table*}

\begin{table*}[!t]
	\setlength{\abovecaptionskip}{0cm}
	\setlength{\belowcaptionskip}{-0.2cm}
	\centering
	\footnotesize
	\caption{Quantitative comparison results (average color error/ruiqa/perceptual scores) of different methods on real benchmarks (Test-R1000, SQUID, UIEB, EUVP, and UFO-120). The top three results are marked in red, blue and green.}
	\renewcommand{\arraystretch}{1.2}
	\begin{tabular}{c|c|c|c|c|c|c|c|c|c|c|c}
		\hline
		\multirow{2}{*}{Methods} &Color Error$\downarrow$ &\multicolumn{5}{c|}{RUIQA$\uparrow$} &\multicolumn{5}{c}{Perceptual Scores$\uparrow$}
		\\\cline{2-12}
		&SQUID &Test-R1000 &SQUID &UIEB &EUVP &UFO-120 &Test-R1000 &SQUID &UIEB &EUVP &UFO-120 \\\hline
		\hline
		\makecell[c]{Fusion-12~\cite{Ancuti2012}} &31.653 &49.204 &44.958 &51.185 &47.375 &50.379 &2.442 &2.049 &2.404 &2.464 &2.462 \\
		\makecell[c]{Fusion-18~\cite{Ancuti2018}} &{\color{red}{6.008}} &{\color{blue}{59.850}} &{\color{blue}{60.430}} &{\color{blue}{60.631}} &{\color{blue}{57.208}} &{\color{blue}{59.499}} &\color{blue}{3.209} &\color{blue}{3.208} &\color{blue}{3.320} &\color{blue}{3.153} &\color{blue}{3.216} \\
		\makecell[c]{HE-Prior~\cite{Li2016UnderwaterIE}} &21.586 &32.287 &23.239 &35.420 &36.549 &34.403 &1.731 &1.072 &1.558 &1.973 &1.580 \\
		\makecell[c]{UIBLA~\cite{Peng2017UnderwaterIR}} &34.214 &42.727 &41.395 &43.170 &40.227 &42.213 &2.020 &1.997 &2.084 &1.891 &2.073 \\
		\makecell[c]{Water-Net~\cite{Li2020AnUI}} &23.352 &45.040 &\color[rgb]{0,0.5,0}{46.793} &\color[rgb]{0,0.5,0}{53.004} &\color[rgb]{0,0.5,0}{50.611} &\color[rgb]{0,0.5,0}{52.976} &2.093 &\color[rgb]{0,0.5,0}{2.740} &2.593 &2.575 &2.618 \\
		\makecell[c]{FUIE-GAN~\cite{islam2020fast}} &26.847 &\color[rgb]{0,0.5,0}{50.784} &46.061 &51.928 &49.150 &52.871 &\color[rgb]{0,0.5,0}{2.522} &2.042 &\color[rgb]{0,0.5,0}{2.782} &\color[rgb]{0,0.5,0}{2.608} &\color[rgb]{0,0.5,0}{2.802} \\
		\makecell[c]{UGAN~\cite{Fabbri2018}} &\color{blue}{10.158} &46.996 &46.461 &48.391 &47.412 &48.218 &1.842 &1.874 &2.031 &2.156 &2.007 \\
		Our TUDA &\color[rgb]{0,0.5,0}{11.718} &{\color{red}{61.436}} &{\color{red}{61.127}} &{\color{red}{60.952}} &{\color{red}{59.262}} &{\color{red}{60.067}} &{\color{red}{3.595}} &{\color{red}{3.993}} &{\color{red}{3.675}} &{\color{red}{3.531}} &{\color{red}{3.537}}\\
		\hline\hline
	\end{tabular}
	\label{table1}
\end{table*}

\begin{figure*}[!t]
	\centering
	\centerline{\includegraphics[scale = 0.546]{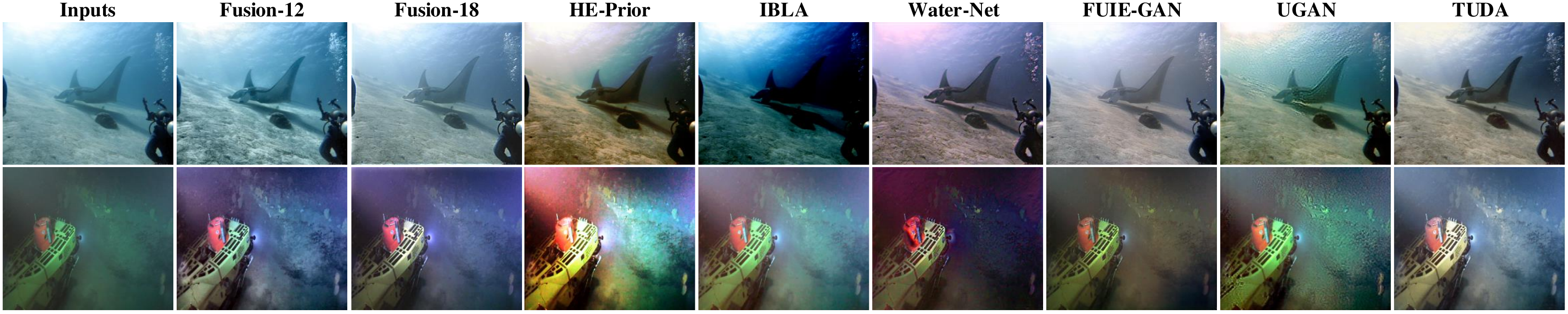}} 
	\caption{Visual comparisons on challenging underwater images sampled from~\textbf{Test-R1000}. From left to right are raw underwater images, and the results of Fusion-12~\cite{Ancuti2012}, Fusion-18~\cite{Ancuti2018}, HE-Prior~\cite{Li2016UnderwaterIE}, UIBLA~\cite{Peng2017UnderwaterIR}, Water-Net~\cite{Li2020AnUI}, FUIE-GAN~\cite{islam2020fast}, UGAN~\cite{Fabbri2018} and our proposed TUDA.}
	\label{Test-1000} 
\end{figure*}

For results on real images, performances are measured by three no-reference underwater quality assessment metrics: UCIQE, UIQM and our proposed RUIQA. For the three metrics, a higher score denotes a better human visual perception. It should be pointed out that UCIQE and UIQM are not sufficient to reflect the performance of various underwater image enhancement methods in some cases~\cite{Li2020AnUI, Li2021}, and thus the scores of UCIQE and UIQM are only for reference. 

In addition, a user study is conducted to more accurately evaluate the visual quality of the results, in which 30 images are randomly selected from each testing dataset to scored. 15 volunteers are invited in this evaluation, and the scoring range is 0 to 5 levels, referring Bad, Poor, Fair, Good and Excellent, respectively. To evaluate the color restoration accuracy of different methods, we also calculate the color restoration accuracy on the average angular reproduction error~\cite{Berman2020UnderwaterSI} on the 16 representative examples presented in the project page\footnote[7]{http://csms.haifa.ac.il/profiles/tTreibitz/datasets/ambient\_forwardlooking/\\index.html} of SQUID. The smaller color error, the better performance.

\subsection{Comparisons with State-of-the-Art Methods}
In this section, we conduct quantitative and visual comparisons on diverse challenging testing dataset. The results of different methods are reported in the following subsections.

\textbf{Quantitative Comparisons:} The quantitative comparison results of different methods on real challenging set are listed in Table~\ref{table10} and Table~\ref{table1}. As presented, HE-Prior achieves the highest scores in term of UCIQE, while our TUDA ranks the fourth best on all challenging set. For the UIQM scores, our method almost achieves the best performance across all datasets, and UGAN ranks the second best. The average values of the color restoration accuracy of different methods on 16 representative examples of SQUID are reported in the second column of Table~\ref{table1}. It can be observed that our TUDA achieves relatively low average error, making a more effectively recovery of a scene’s colors. Fusion-18 obtains the lowest color error than other methods. However, its performance is not as good as our TUDA in terms of RUIQA and Perceptual Scores. Among them, our TUDA achieves the best performance. Such results demonstrate that our TUDA produces visually more convincing results and has more robust performance in handling images taken in a diversity of underwater scenes. 

\begin{figure*}[!t]
	\centering
	\centerline{\includegraphics[scale = 0.546]{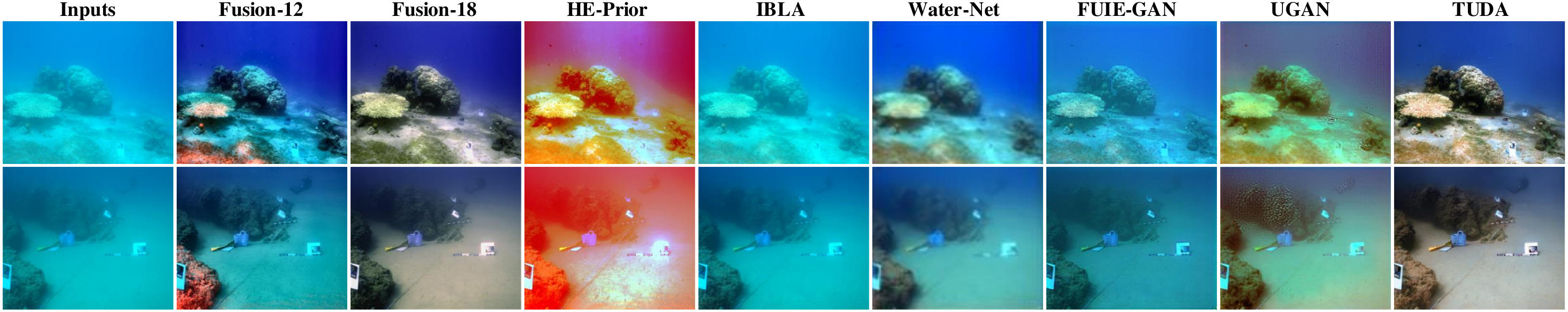}} 
	\caption{Visual comparisons on challenging underwater images sampled from~\textbf{SQUID}. From left to right are raw underwater images, and the results of Fusion-12~\cite{Ancuti2012}, Fusion-18~\cite{Ancuti2018}, HE-Prior~\cite{Li2016UnderwaterIE}, UIBLA~\cite{Peng2017UnderwaterIR}, Water-Net~\cite{Li2020AnUI}, FUIE-GAN~\cite{islam2020fast}, UGAN~\cite{Fabbri2018} and our proposed TUDA.}
	\label{SQUID} 
\end{figure*}

\begin{figure*}[!t]
	\centering
	\centerline{\includegraphics[scale = 0.546]{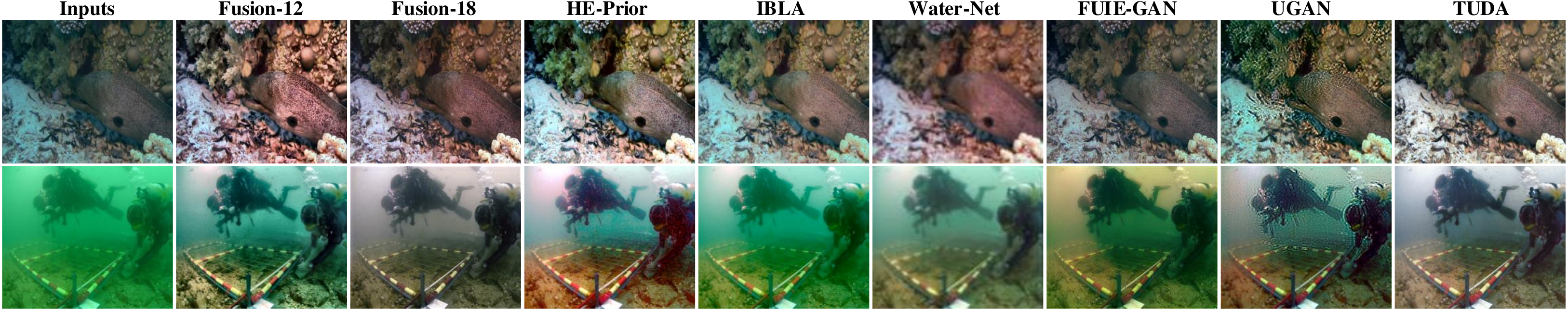}} 
	\caption{Visual comparisons on challenging underwater images sampled from~\textbf{UIEB}. From left to right are raw underwater images, and the results of Fusion-12~\cite{Ancuti2012}, Fusion-18~\cite{Ancuti2018}, HE-Prior~\cite{Li2016UnderwaterIE}, UIBLA~\cite{Peng2017UnderwaterIR}, Water-Net~\cite{Li2020AnUI}, FUIE-GAN~\cite{islam2020fast}, UGAN~\cite{Fabbri2018} and our proposed TUDA.}
	\label{UIEB} 
\end{figure*}

\begin{figure*}[!t]
	\centering
	\centerline{\includegraphics[scale = 0.546]{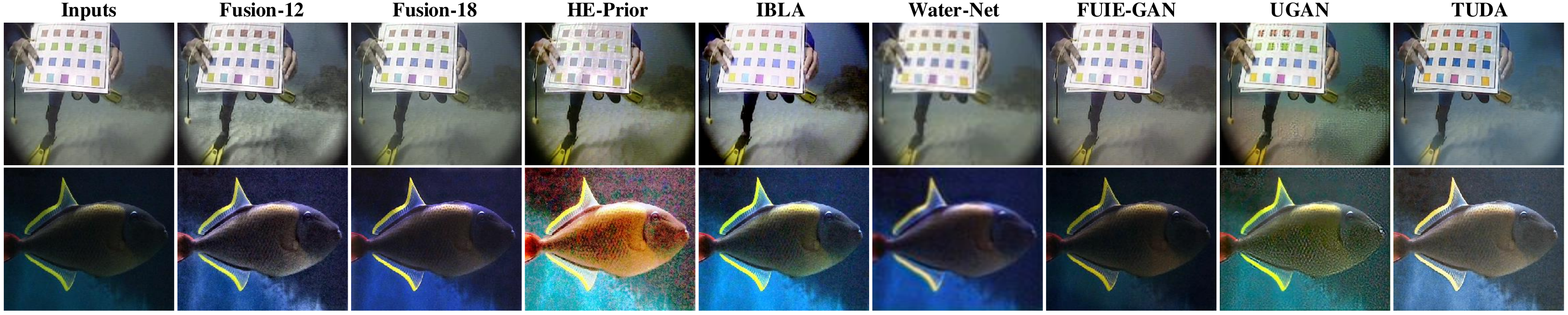}} 
	\caption{Visual comparisons on challenging underwater images sampled from~\textbf{EUVP}. From left to right are raw underwater images, and the results of Fusion-12~\cite{Ancuti2012}, Fusion-18~\cite{Ancuti2018}, HE-Prior~\cite{Li2016UnderwaterIE}, UIBLA~\cite{Peng2017UnderwaterIR}, Water-Net~\cite{Li2020AnUI}, FUIE-GAN~\cite{islam2020fast}, UGAN~\cite{Fabbri2018} and our proposed TUDA.}
	\label{EUVP} 
\end{figure*}

\begin{figure*}[!t]
	\centering
	\centerline{\includegraphics[scale = 0.546]{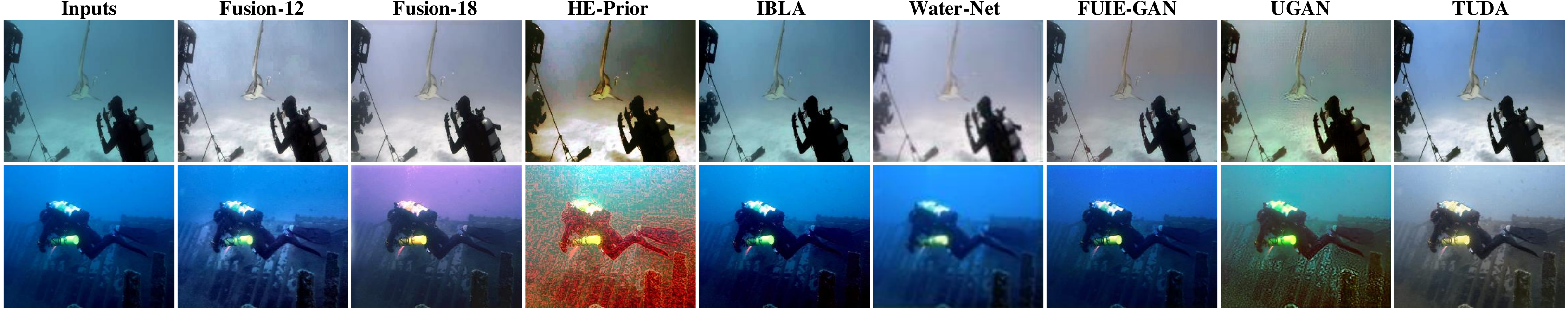}} 
	\caption{Visual comparisons on challenging underwater images sampled from~\textbf{UF0-120}. From left to right are raw underwater images, and the results of Fusion-12~\cite{Ancuti2012}, Fusion-18~\cite{Ancuti2018}, HE-Prior~\cite{Li2016UnderwaterIE}, UIBLA~\cite{Peng2017UnderwaterIR}, Water-Net~\cite{Li2020AnUI}, FUIE-GAN~\cite{islam2020fast}, UGAN~\cite{Fabbri2018} and our proposed TUDA.}
	\label{UFO-120} 
\end{figure*}

The deep methods trained based on real data including Water-Net and FUIE-GAN perform relatively well, but they do not restore green or some excessively distorted images due to ignoring the intra-domain gap among real underwater images itself, and thus its performance is limited. UGAN trains the model using the synthetic data generated by GAN methods. Since the inter-domain gap is not effectively reduced, the results often contain various artifacts, and thus the subjective effect is not good, and the score is relatively low.

There is an interesting finding from the quantitative results. He-Prior almost achieves the highest UCIQE scores on all real datasets. However, its perceptual score is the lowest, which means it has the worst subjective quality. In our opinion, this is mainly due to the fact that UCIQE pays too much attention to local features (color) and ignores the entire image, and thus it is not consistent with human visual perception in some cases, especially when the enhanced image is over-enhanced (please refer to Fig.\ref{IQA_test})~\cite{Li2020AnUI}.

\textbf{Visual Comparisons:} Some visual comparisons on~\textbf{Test-R1000} and~\textbf{SQUID} are shown in Fig.\ref{Test-1000} and Fig.\ref{SQUID}, respectively. For these images, except for~Fusion-18~\cite{Ancuti2018} and our TUDA, other competing methods cannot achieve satisfactory results. Some of them even introduce undesirable color artifacts in their enhanced results to some extend, such as~Fusion-12~\cite{Ancuti2012},~HE-Prior~\cite{Li2016UnderwaterIE}, FUIE-GAN~\cite{islam2020fast} and~UIBLA~\cite{Peng2017UnderwaterIR}. Meanwhile, most methods fail to restore the structural details of underwater scene, in which UGAN~\cite{Fabbri2018} even introduces serious artifacts at the boundary of objects. Fusion-18~\cite{Ancuti2018} can restore better color than other methods, but the performance in recovering object details is not as good as our TUDA.

The results of different methods on challenging underwater images sampled from~\textbf{UIEB} and~\textbf{EUVP} are presented in Fig.\ref{UIEB} and Fig.\ref{EUVP}. As presented, for the image with the greenish tone, our TUDA significant removes the haze and color casts, and effectively recovers details, producing visually pleasing results. In comparison, all the comparison methods cannot produce the realistic color. Most of them suffer obvious over-enhancement and under-enhancement, such as~Fusion-12~\cite{Ancuti2012},~Fusion-18~\cite{Ancuti2018},~HE-Prior~\cite{Li2016UnderwaterIE},~UIBLA~\cite{Peng2017UnderwaterIR} and Water-Net~\cite{Li2020AnUI}.~Fusion-18~\cite{Ancuti2018} even lost the original color of the object in the second image. For these low-light underwater images, most methods generate unrealistic results with color artifacts and cannot effectively improve the visibility of objects, and often amplify noise in their enhanced results. Our TUDA not only can effectively increase the brightness of images but also refine the object edges, produce realistic results with correct color from extremely noisy. 

Visual comparisons on challenging underwater images sampled from~\textbf{UFO-120} are shown in~Fig.\ref{UFO-120}. Compared to most existing methods, our TUDA significantly reduces color distortion and satisfactorily removes blurriness. It can be seen that the images enhanced by~HE-Prior~\cite{Li2016UnderwaterIE} have obvious reddish color shift and artifacts in some regions. Besides, UGAN~\cite{Fabbri2018} often introduces undesirable artifacts at the boundary of objects. Most methods cannot correct the colors well, even amplify color deviation (e.g., the color of background).~Fusion-18~\cite{Ancuti2018} can produce relatively good results. However, they still contain numerous noises and color distortion. 

All the above quantitative and visual comparison results demonstrate that considering both reduce the inter-domain gap and the intra-domain gap in our TUDA can produce visually pleasing results and have more robust performance. Due to the limited space, more experimental results are given in the supplementary material.

\subsection{Ablation Studies and Analysis}
In this section, we first evaluate the performance of our proposed RUIQA and analyze its superiority. Subsequently, a series of ablation studies are conducted to analyze the contribution of each proposed component. In addition, we study the influence of different ratio of $\lambda$ on intra-domain adaptation training and TUDA testing.

\subsubsection{Effectiveness of the Proposed RUIQA method}
\begin{figure}[!t]
	\centering
	\centerline{\includegraphics[width=8.6cm, height=5 cm]{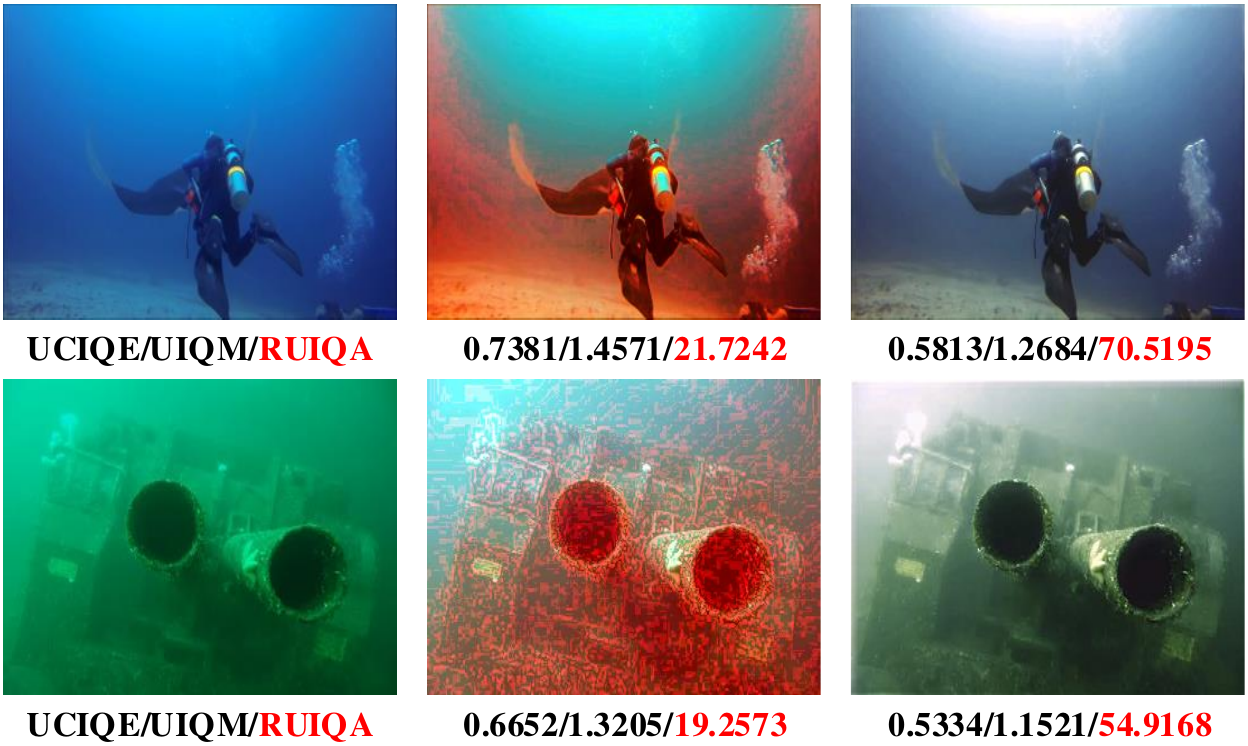}}
	\caption{Visual comparisons in terms of UCIQE, UIQM and our proposed RUIQA metrics. It is obvious that our quantitative scores can better represent subjectively quality.}
	\label{IQA_test}
\end{figure}

As mentioned above, $90 \%$ image pairs (i.e., 7200) of the underwater ranking data are randomly selected as training data, and the other $10 \%$ image pairs (i.e., 800) are used for IQA testing. To validate the generalization ability of our RUIQA, we compare it with two state-of-the-art methods: UCIQE and UIQM in terms of two metrics: Spearman Rank Order Correlation Coefficient (SROCC) and Pearson’s linear correlation coefficient (PLCC). Table~\ref{table2} lists the corresponding comparison results. As shown, our method achieves the best performance, even has good correlation with MOS on the order of 0.900 and achieves the gain of 0.5 to 0.65 in comparison to UCIQE and UIQM, showing the superiority of our RUIQA metric. A visual comparison is also shown in Fig.\ref{IQA_test}. The larger values indicate a better perceptual quality. Obviously, our RUIQA can more accurately reflect the perceptual quality of the image.

In addition, an ablation study is conducted to analyze the contribution of each individual component using the following settings: 1) UIQA: using the IQA network to directly predict image quality score; 2) PUIQA: using the ResNet50 network pre-trained on ImageNet data as our initialization backbone model; 3) RUIQA: using the ResNet50 network pre-trained on our rank data as our initialization backbone model. As presented in table~\ref{table2}, we can see that our RUIQA achieves the best evaluation performance and is significantly better than UIQA and PUIQA. It's worth mentioning that the ImageNet has more than 1.28 million images and our rank training dataset only contains 720 image pairs. This indicates that the pre-trained ResNet50 network on the rank dataset can capture sufficient perceptual quality information of underwater image, and then quickly help the IQA task better predict the image quality score.

\begin{table}
	\setlength{\abovecaptionskip}{0cm}
	\setlength{\belowcaptionskip}{-0.2cm}
	\centering
	\footnotesize
	\caption{SROCC and PLCC results of different methods and ablation studies on the rank test set.}
	\renewcommand{\arraystretch}{1.2}
	\begin{tabular}{c|c|c|c|c|c}
		\hline\hline
		Method     & UCIQE   & UIQM &UIQA &PUIQA &RUIQA \\ \hline
		SROCC $\uparrow$ &0.245 &0.400 &0.796 &0.835 &{\color{red}{0.900}}\\ \hline
		PLCC $\uparrow$ &0.326 &0.422 &0.826 &0.862 &{\color{red}{0.904}}\\ 
		\hline\hline
	\end{tabular}
	\label{table2}
\end{table}

\begin{table}[!t]
	\setlength{\abovecaptionskip}{0cm}
	\setlength{\belowcaptionskip}{-0.2cm}
	\centering
	\footnotesize
	\caption{The ablation study of the proposed inter-domain adaptation module on the synthetic test set and the real test Test-R1000}
	\renewcommand{\arraystretch}{1.2}
	\begin{tabular}{c|c|c|c|c}
		\hline\hline
		Methods &\multicolumn{2}{c|}{Test-S1800$\uparrow$} &\multicolumn{2}{c}{Test-R1000$\uparrow$}\\\hline
		&PSNR$\uparrow$ &SSIM$\uparrow$ &RUIQA$\uparrow$ &Perceptual Score$\uparrow$\\\hline
		\makecell[c]{BL} &{\color{red}{27.284}} &{\color{red}{0.949}} &55.534 &3.146\\
		\makecell[c]{BL+ITE} &27.065 &0.947 &{\color{red}{59.250}} &\color{red}{3.390}\\
		\hline\hline
	\end{tabular}
	\label{table3}
\end{table}

\begin{table}[!t]
	\setlength{\abovecaptionskip}{0cm}
	\setlength{\belowcaptionskip}{-0.2cm}
	\centering
	\footnotesize
	\caption{{The ablation study of the proposed intra-domain adaptation module on the real test set Test-R1000.}}
	\renewcommand{\arraystretch}{1.4}
	\begin{tabular}{c|c|c}
		\hline
		Method & RUIQA $\uparrow$ &Perceptual Score $\uparrow$ \\ \hline
		BL + ITE & 59.250 & 3.390  \\ \hline
		BL + ITE + ITA & {\color{red}{61.436}} &\color{red}{3.595} \\ \hline
	\end{tabular}
	\label{table4}
\end{table}

\subsubsection{Effectiveness of the inter-domain adaptation phase}
\begin{figure}[!t]
	\centering
	\centerline{\includegraphics[width=8.6cm, height=6.8cm]{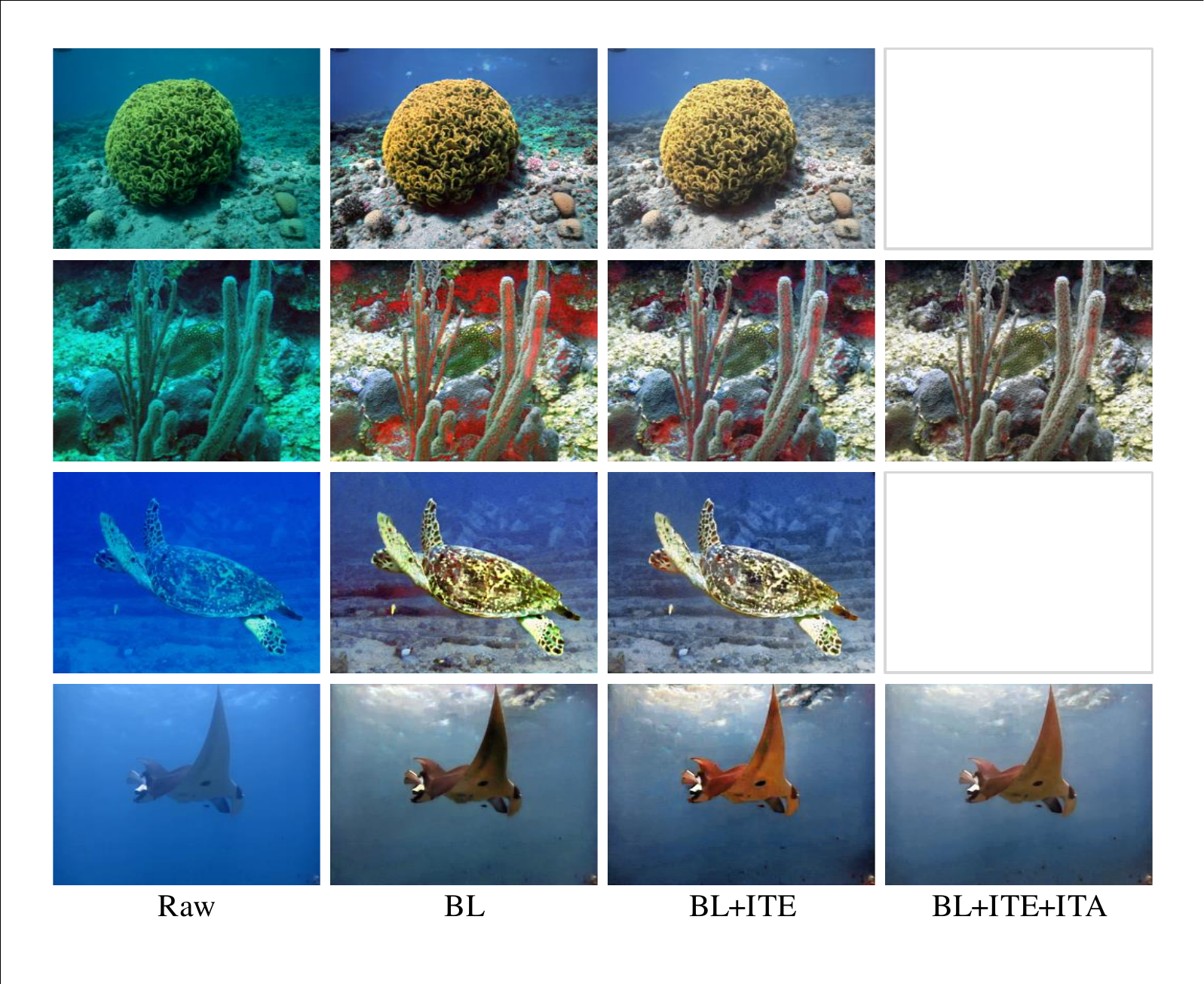}}
	\caption{Visual comparisons on some easy and hard samples. We can clearly see that our method can effectively handle easy and hard samples, especially on hard samples, our full model generates the most visually pleasing results.}
	\label{Compare}
\end{figure}

In this part, we perform an ablation study of 60 images randomly chosen from enhanced images in~\textbf{Test-R1000} to evaluate the effectiveness of the inter-domain adaptation, as follows: 1) BL: baseline network (trained on synthetic data); 2) BL+ ITE: baseline network with the inter-domain adaptation, i.e., dual-alignment network. Results are listed in Table~\ref{table3}. It can be seen that baseline network has only slightly higher PSNR values in comparison to our dual-alignment network (0.219dB higher on average), but the perceptual quality is far worse than the dual-alignment network (3.716 and 0.244 lower on average in RUIQA and perceptual score of user study, respectively). Such results show that our inter-domain adaptation part generates the enhanced results with well reconstructed details (high fidelity) and good perceptual visual quality. Some examples are shown in Fig.\ref{Compare}, in which our inter-domain adaptation phase can better correct color casts and avoid over-enhancement than baseline network.

\begin{table}!t]
	\setlength{\abovecaptionskip}{0cm}
	\setlength{\belowcaptionskip}{-0.2cm}
	\centering
	\footnotesize
	\caption{{The ablation study on hyper-parameter $\lambda$ for dividing the real underwater data into the easy and hard samples.}}
	\renewcommand{\arraystretch}{1.4}
	\begin{tabular}{c|c|c|c|c|c|c}
		\hline
		$\lambda$ &0.40 &0.45& 0.50& 0.55& 0.60& 0.65 \\ \hline
		RUIQA$\uparrow$ & {\color{red}{62.03}} & 61.39 & 61.78 & 61.31 & 61.20 & 61.39  \\ \hline
	\end{tabular}
	\label{table5}
\end{table}

\begin{table}[!t]
	\setlength{\abovecaptionskip}{0cm}
	\setlength{\belowcaptionskip}{-0.2cm}
	\centering
	\footnotesize
	\caption{The Parameters, Flops and running time of different learning-based methods.}
	\renewcommand{\arraystretch}{1.2}
	\begin{tabular}{c|c|c|c|c}
		\hline
		& Our TUDA &UGAN &FUIE-GAN &Water-Net \\ \hline
		Flops (G) & 174.4 &3887 & 0.008 & 1937  \\ \hline
		Parameters (M) & 31.36 &57.17 & 4.216 & 1.091  \\ \hline
		Running time (s) &0.051&0.009 &0.083 & 0.582 \\
		\hline
	\end{tabular}
	\label{table11}
\end{table}
 
\subsubsection{Effectiveness of the intra-domain adaptation phase}
In the intra-domain adaptation part, we conduct an ablation study of 60 images randomly selected from enhanced results in~\textbf{Test-R1000} with the following settings: 1) BL+ITE: baseline network with the inter-domain adaptation; 2) BL+ITE+ITA: baseline network with the inter-domain and the intra-domain adaption. The averaged RUIQA value and the averaged perceptual score are reported in Table~\ref{table4}. It can be seen BL+ITE+ITA achieves better performance, even the average performance gains up to 2.186 and 0.205 in two metrics, respectively. This indicates that intra-domain adaptation can effectively process hard samples and significantly improve the perceptual quality of the image, making enhanced results are more subject to human preferences. In addition, a few samples are illustrated in Fig.\ref{Compare}. It can be noted that if only conduct inter-domain adaptation phase, the results of some hard samples still contain some noises and over-enhancement artifacts in some region. In other words, our intra-domain adaptation part is robust for real-world extremely hard underwater image enhancement, producing visually more pleasing results.

\subsubsection{Analysis of Hyperparameter $\lambda$}
$x_{n}$ denotes the real underwater images for inter-domain training, in our work, $\mathrm{n}\subset(1,2900)$. The inter-domain enhancement part $G_{{inter}}^{E}$ receives the input $x_n$ and outputs the enhanced image $\widehat{x_{n}}$. The proposed RUIQA is used to evaluate their perceptual quality score, i.e.,~$\operatorname{MOS}_{n}=\operatorname{RUIQA}\left(\widehat{x_{n}}\right)$. We rank the $\operatorname{MOS}_{n}$ value (i.e., $\operatorname{MOS}_{n}^{{Rank}}$) and select the corresponding $\operatorname{MOS}_{n}^{{Rank}}$ value of the ratio $\lambda$ as the threshold (i.e., $\operatorname{MOS}_{n}^{\lambda * \operatorname{Rank}}$) to separate the real underwater data $x_n$ into easy and hard samples (i.e., $x_e$ and $x_h$) for intra-domain training and the whole framework testing. Thus, different values of the ratio $\lambda$ will have a significant impact on subsequent operations. Here, some experiments are conducted to decide the optimal $\lambda$ in our framework. For a selected ratio $\lambda$, we first conduct intra-domain adaptation training. Then, the 2900 real training data (\textbf{Test-S2900}) is used as validate data in the test pipeline (see Fig.\ref{test}). Finally, we predict the average perceptual quality score of 2900 enhanced images in term of the RUIQA metric, and set it as the metric for selecting $\lambda$. Results are shown in Table~\ref{table5}, where it can be observed that $\lambda=0.4$, the proposed method can achieve better performance.
\label{section:me2}

\subsection{Model Complexity Analysis}
We compare the flops, parameters and time cost of different learning-based methods on a PC with an Intel(R) i5-10500 CPU, 16.0GB RAM, and a NVIDIA GeForce RTX 2080 Super. The test dataset is UIEB benchmark, which includes 950 images and its size is 256x256x3. The source codes and test parameters of all the compared methods are provided by their authors, and the results are presented in Table~\ref{table11}.

As presented,  the computational aspect and time cost of our method are ideal. UGAN has the shortest running time, but its flops and parameters are the most, far exceeding our method. The size, computation and time cost of FUIE-GAN are less than our TUDA. However, the generalization performance on four real underwater benchmarks is limited, not as good as our method. The parameters of Water-Net is the least, but its flops and time cost are large. Such results demonstrate that our TUDA can achieve good performance and efficiency.

\section{CONCLUSION}
In this paper, a novel two-phase underwater domain adaptation method is proposed for enhancing underwater images, which contains an inter-domain adaptation and an intra-domain adaptation phase to jointly optimize the inter-domain gap and the inter-domain gap. Firstly, a dual-alignment network is introduced to jointly perform image-level and feature-level alignment using adversarial learning for better closing the inter-domain gap. Secondly, a simple yet efficient rank-based underwater IQA method is developed, which can evaluate the perceptual quality of underwater images with the aid of rank information, named RUIQA. Finally, coupled with the proposed RUIQA, an easy/hard adaptation technique is conducted to effectively reduce the intra-domain gap between easy and hard samples. Extensive experiments on four real underwater benchmarks demonstrate that our TUDA can significantly perform favorably against other state-of-the-art algorithms, particularly on eliminating color deviation, increasing contrast and avoiding over-enhancement.

\bibliographystyle{ieeetr}
\bibliography{TUDA}
\end{document}